\documentclass{article}

\usepackage[preprint]{neurips_2025}

\usepackage[utf8]{inputenc} %
\usepackage[T1]{fontenc}    %
\usepackage{hyperref}       %
\usepackage{url}            %
\usepackage{booktabs}       %
\usepackage{amsfonts}       %
\usepackage{nicefrac}       %
\usepackage{microtype}      %
\usepackage{xcolor}         %

\usepackage{color}
\usepackage{colortbl}
\usepackage{url}
\usepackage{booktabs}
\usepackage{multirow}
\usepackage{makecell}
\usepackage{siunitx}
\usepackage{mathtools}
\usepackage{bm}
\usepackage{bbm}
\usepackage{amsmath}
\usepackage{amsthm}
\usepackage{amssymb}
\usepackage{algorithmic}
\usepackage{graphicx}
\usepackage{caption}
\usepackage{subcaption}
\usepackage{wrapfig}
\usepackage{appendix}
\usepackage{xspace}

\makeatletter

\theoremstyle{plain}

\theoremstyle{remark}
\newtheorem{remark}{\protect\remarkname}

\allowdisplaybreaks

\newcommand{\R}{{\mathbb{R}}}

\newcommand{\bk}{\bm{k}}

\newcommand{\bo}{\bm{o}}

\newcommand{\bq}{\bm{q}}

\newcommand{\bs}{\bm{s}}

\newcommand{\bv}{\bm{v}}

\newcommand{\bx}{\bm{x}}

\newcommand{\bz}{\bm{z}}

\newcommand{\balpha}{\bm{\alpha}}

\newcommand{\softmax}{\mathsf{Softmax}}
\newcommand{\concat}{\mathsf{Concat}}

\newcommand{\trans}{\mathsf{T}}
\newcommand{\V}{\mathbb{V}}
\newcommand{\Tcal}{\mathcal{T}}

\newcommand{\Rcal}{\mathcal{R}}
\newcommand{\Ical}{\mathcal{I}}

\newcommand{\comma}[1]{\texttt{Comma\_#1}\xspace}
\newcommand{\start}[1]{\texttt{Start\_#1}\xspace}
\newcommand{\eend}[1]{\texttt{End\_#1}\xspace}
\newcommand{\freq}[2]{\texttt{#1\_#2}\xspace}
\newcommand{\fs}{\texttt{FS}\xspace}

\newcommand{\append}{\texttt{APPEND}\xspace}
\newcommand{\prepend}{\texttt{PREPEND}\xspace}

\newcommand{\bos}{\texttt{\textless BOS\textgreater}\xspace}
\newcommand{\eos}{\texttt{\textless EOS\textgreater}\xspace}
\newcommand{\pause}{\texttt{\textless PAUSE\textgreater}\xspace}

\newcommand{\llamas}{Llama-3.2-1B\xspace}
\newcommand{\llamal}{Llama-3.1-8B\xspace}

\newcommand{\query}{\textbf{\textsc{Query:}}\,}
\newcommand{\response}{\textbf{\textsc{Response:}}\,}

\makeatother

\providecommand{\remarkname}{Remark}

\usepackage{enumitem}

\title{Enhancing Latent Computation in Transformers\\with Latent Tokens}

\author{%
Yuchang Sun, Yanxi Chen, Yaliang Li, Bolin Ding \\
Alibaba Group \\
\texttt{\{sunyuchang.syc, chenyanxi.cyx, yaliang.li, bolin.ding\}@alibaba-inc.com}
}

\begin{document}

\maketitle

\begin{abstract}
  Augmenting large language models (LLMs) with auxiliary tokens has emerged as a promising strategy for enhancing model performance. In this work, we introduce a lightweight method termed \emph{latent tokens}; these are dummy tokens that may be non-interpretable in natural language but steer the autoregressive decoding process of a Transformer-based LLM via the attention mechanism. The proposed latent tokens can be seamlessly integrated with a pre-trained Transformer, trained in a parameter-efficient manner, and applied flexibly at inference time, while adding minimal complexity overhead to the existing infrastructure of standard Transformers. We propose several hypotheses about the underlying mechanisms of latent tokens and design synthetic tasks accordingly to verify them. Numerical results confirm that the proposed method noticeably outperforms the baselines, particularly in the out-of-distribution generalization scenarios, highlighting its potential in improving the adaptability of LLMs.
\end{abstract}

\section{Introduction}
Transformer-based large language models (LLMs) such as Llama~\cite{meta2024llama3herdmodels} and GPT~\cite{openai2024gpt4} generate a response for any given query 
via next-token prediction. 
In this process, LLMs generate tokens one after the other in immediate succession, which often proves insufficient for solving complex tasks~\cite{faith,pitfalls,william2024expressive}. 
Recent works have been exploring various approaches to augmenting LLMs with auxiliary tokens inserted into the input sequence to provide intermediate computations. These tokens can either carry explicit semantic meanings (e.g., Chain-of-Thought (CoT)~\cite{cot}, scratchpad~\cite{scratchpad}, and LLM prompting~\cite{BrownMRSKDNSSAA20,yan2024understanding}) or serve as abstract placeholders without direct linguistic interpretation~\cite{google2024think,pfau2024lets,coconut}.
Specifically, the Transformer-based LLM~\cite{vaswani2017transformer} is allowed to conduct the forward passes for some additional tokens before generating the subsequent tokens, thereby improving the quality of the generated outputs.
Compared with ``interpretable'' approaches~\cite{cot,scratchpad}, ``non-interpretable'' methods insert some special tokens into the input sequence to provide additional computation, which may not have explicit meanings themselves. This type of method, such as pause tokens~\cite{google2024think}, filler tokens~\cite{pfau2024lets}, and many others, shows potential in improving model performance beyond the limitations of natural language.\looseness=-1

Existing methods with non-interpretable tokens typically insert the tokens at fixed positions, including the start of the query~\cite{prompt} or the end of the query~\cite{google2024think,coconut}. This limits their effectiveness in two critical aspects: (1) the effect of these tokens diminishes progressively with increasing sequence length, leading to possible failure in the long-generation scenarios; (2) they struggle to generalize to longer, out-of-distribution (OOD) sequences beyond the training distribution. These limitations highlight the need for a principled framework that can effectively coordinate latent computation while preserving the generalization abilities.

\begin{figure*}[!t]
    \centering
    \includegraphics[width=\textwidth]{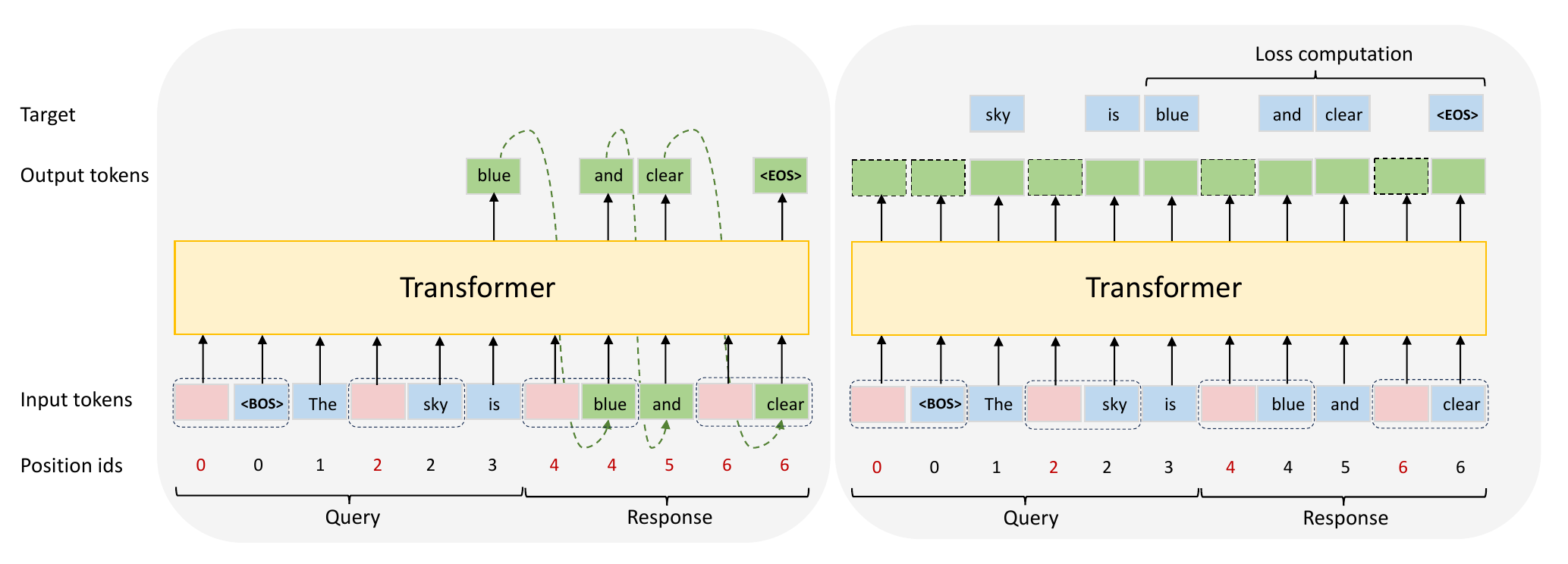}
    \caption{A decoder-only Transformer model with latent tokens. 
    \textbf{Left:} The inference process. 
    \textbf{Right:} The training process.
    In this visualization, we assume that one latent token is inserted periodically, once every two verbal tokens, into the original sequence, though it can be much more general, as discussed in Section~\ref{sec:design_choices}.
    }
    \label{fig:latent_token}
\end{figure*}

To this end, we propose a unified method, designed to be simple yet effective, of augmenting any decoder-only Transformer with \emph{latent tokens}.
These are dummy tokens that may not have a correspondence in natural language, can be applied flexibly at inference time,
and enhance latent computation before the prediction of any output token.
We present a detailed design of latent tokens, including the inference and training processes. To ensure general applicability, we carefully design the positional encoding and loss function for latent tokens, among other aspects. With these designs, latent tokens are broadly applicable for general tasks without relying on heavy training of the full Transformer, and fully compatible with existing infrastructure for standard Transformers.
Despite its simplicity, our approach demonstrates consistent improvements in LLM performance across several benchmark tasks (see Appendix~\ref{sec:benchmark}).

While the results are encouraging, it remains unclear which tasks benefit most from latent tokens and the underlying reasons for their effectiveness.
In this work, we conduct an in-depth investigation to shed light on the roles that latent tokens play during the generation process. 
Concretely, we propose several hypotheses about their utilities: (i)~\emph{self-prompting for long and consistent generation}, (ii)~\emph{assisting information retrieval}, and (iii)~\emph{improving instruction adherence in longer OOD sequences}.
For each hypothesis, we design a corresponding synthetic task to evaluate the performance gap between our method and existing approaches.
The observed performance improvements --- particularly in the out-of-distribution scenarios --- highlight the benefits and suggest possible underlying mechanisms of the latent tokens.

\section{Preliminaries for Standard Transformers}

Let $\bs = \{s_1, \dots, s_t, \dots, s_n \}$ denote a sequence of tokens with position indices (or IDs) $\{1, \dots, t, \dots, n\}$.
A {decoder-only} Transformer~\cite{vaswani2017transformer} model with parameters $\theta$ predicts tokens one at a time autoregressively.
That is, each output token $s_{t+1}$ is generated based on the input tokens $\bs_{1:t} = \{s_1, \dots, s_{t}\}$ in front of it.
In a typical fine-tuning scenario,
a sequence $\bs$ in the training data consists of a query and the corresponding response.
The standard objective for training or fine-tuning is the negative log-likelihood of the target tokens for each sample $\bs$ in training set $\Tcal$, which is given by:\looseness=-1
\begin{equation}
    \min_{\theta} \, \sum_{\bs\in\Tcal} \sum_{t\in \Rcal_{\bs}} -\log p_{\theta}(s_{t+1} | \bs_{1:t}),
\end{equation}
where $\Rcal_{\bs}$ denotes the token indices of the response in sample sequence $\bs$, and $p_{\theta}$ is the probability distribution of the output token given model $\theta$ and context $\bs_{1:t}$.

At inference time, a query sequence is fed into the Transformer model, and the model generates a response in a token-by-token manner. For example, using the greedy decoding strategy, the model employs the token with the highest prediction probability as the next token: 
\begin{equation}
    s_{t+1} = \arg\max_{s\in\V} \, p_{\theta}(s | \bs_{1:t}),
    \label{eq:greedy}
\end{equation}
where $\V$ denotes the vocabulary set.
The predicted token is then fed back to the model as part of the context $\bs_{1:t+1}$ to generate the next token $s_{t+2}$, which is repeated until the ending signal or the maximum number of tokens is met.

At the token level, the prediction of the next token with the standard Transformer only involves the forward computation of one token.
The expressive power of the model can be restricted by the limited computation, resulting in unsatisfactory performance in some tasks requiring complex reasoning or long-term dependencies~\cite{william2024expressive,NowakSBC24,google2024think,pfau2024lets}.
To this end, in the next section, we propose a method of adding latent tokens during sequence generation to provide additional latent computation at inference time.\looseness=-1

\section{Latent Tokens: A Unified Methodology}\label{sec:method}

In this section, we introduce our approach to incorporating \emph{latent tokens} into the decoder-only Transformer, as shown in Fig.~\ref{fig:latent_token}. These latent tokens offer some additional computation to assist the model in generating the output tokens via the attention mechanism. However, they are not expected to generate extra outputs with explicit verbal meaning.
There are various desired properties of latent tokens that we strive to achieve, for example:
\begin{itemize}[left=5pt]
    \item \textbf{General applicability:} The latent tokens can be incorporated into any decoder-only Transformer, and the methodology should be compatible with different types of tasks, rather than limited to some specific ones.
    \item \textbf{Latent computation:} The latent tokens are used to assist the model in generating the output tokens, while they may not be intended to produce extra interpretable sequences themselves.
    \item \textbf{Minimal disturbance:} When introducing latent tokens to a pre-trained Transformer, 
    the negative disturbance on the original distribution of verbal tokens\footnote{We use ``verbal tokens'' to distinguish the original tokens of a standard Transformer from the latent tokens.} should be minimized.
\end{itemize}

On a high level, a group of $m$ latent tokens $u_1, \dots, u_m$ is introduced as new tokens that are not present in the original vocabulary set. Each latent token $u_i$ with index $i$ corresponds to a learnable vector $\bz_i\in \R^{d}$, where $d$ stands for the embedding dimension. In the following, we present the detailed design of the proposed method, including the inference process, the training process, and the design choices of the latent tokens.

\subsection{Inference with Latent Tokens}
At inference time, the latent tokens are incorporated in both the query and generated response to provide additional computation as assistance.
For the query sequence $\{s_1, \dots, s_t\}$, we assume that some latent tokens have been added at proper positions beforehand, following the same procedure as in the optimization process (to be explained in Sections~\ref{sec:optimize_latent} and~\ref{sec:design_choices}). Based on the query with inserted latent tokens, the Transformer generates the first token $s_{t+1}$ of the response.
Note that the output space of predicted tokens is still the original vocabulary, as we do not expect the model to generate the latent tokens in the output sequence.

During the autoregressive generation process, inserting latent tokens into the sequence also helps the prediction of subsequent tokens. 
To enable this design, we need to allocate the proper position IDs to the generated verbal tokens and latent tokens (if any) such that the Transformer accurately captures their relative position information.
Specifically, we assign the position ID $t+1$ to the generated token $s_{t+1}$. Assuming that a group of latent tokens $u_1, \dots,u_m$ are intended to help $s_{t+1}$ in generating subsequent tokens, we prepend them before token $s_{t+1}$ on the input side of the model\footnote{Another choice, i.e., appending latent tokens after a verbal token, is discussed in Appendix~\ref{sec:append_or_prepend}.}.
The sequence, now augmented with the newly added latent tokens and expressed as $\{s_1, \dots, s_t, u_1, \dots,u_m, s_{t+1} \}$, 
is then fed back to the Transformer for next-token prediction.

\begin{figure}[!t]
    \centering
    \includegraphics[width=0.65\linewidth]{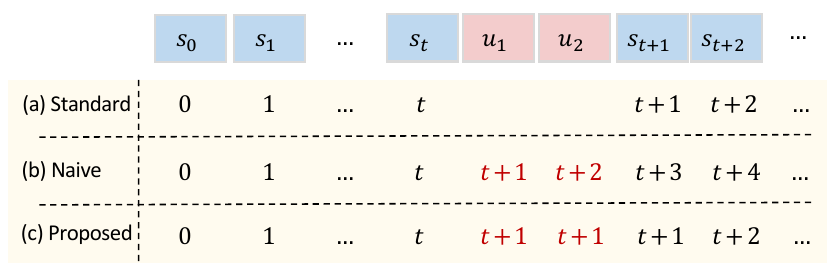}
    \caption{Comparisons of position IDs (a) with the standard Transformer, (b) with a naive way of inserting latent tokens, and (c) with the proposed method.}
    \label{fig:position_IDs}
\end{figure}

\paragraph{Position encoding of latent tokens.}
The position information of input tokens is crucial for revealing the structure of the input sequence~\cite{ke2021rethinking,zhao2024length}.
When incorporating latent tokens into the input sequence, it is important to ensure that they do not disturb the position information of the original input tokens. To achieve this goal, we propose to use the same position encoding for the latent tokens as their following verbal tokens, as shown in Fig.~\ref{fig:position_IDs}.
To be specific, we keep the position IDs of all verbal tokens unchanged and assign position ID $t+1$ to the group of latent tokens located right before verbal token $s_{t+1}$.
This design ensures that the model still captures the position information of the verbal tokens and the latent tokens, which is crucial for the model to understand the input sequence properly.
It is also worth noting that such a design is compatible with essentially \emph{any} scheme that incorporates positional information based on position IDs of tokens, ranging from absolute position embedding in the original Transformer~\cite{vaswani2017transformer} to more recent approaches like Rotary Position
Embedding~\cite{su2024roformer} and more.
In Appendix~\ref{sec:position_encoding}, we discuss the rationale and benefits of the above position encoding scheme more formally.

\subsection{Optimizing Latent Tokens}\label{sec:optimize_latent}

To optimize the latent tokens, we pre-process the training data by inserting the latent tokens into the input sequences. In principle, we can insert latent tokens at any position of the input sequences.
Assuming that a group of $m$ latent tokens is used between two input tokens $s_{t}$ and $s_{t+1}$, the model is optimized to predict the output tokens based on the augmented input sequence $\hat{\bs} = \{s_1, \dots, s_t, u_1, \dots, u_{m}, s_{t+1}, \dots, s_n\}$ in dataset $\hat{\Tcal}$.
For the current work, this optimization process is restricted to the fine-tuning stage of a pre-trained Transformer model, while the pre-training stage remains unchanged. In the following, we focus on optimizing the latent tokens alone while freezing the model parameters of the pre-trained Transformer.
Compared with existing works~\cite{google2024think,pfau2024lets,coconut}, this method provides a more lightweight way to enable latent computation.

\paragraph{Loss function design.}
The latent tokens are expected to assist the context sequence $\hat{\bs}_{1:k}$ in predicting the next token $\hat{s}_{k+1}$, which is expressed as:
\begin{equation}
    \min_{\bz} \, \sum_{\hat{\bs}\in\hat{\Tcal}} \sum_{k\in{\Ical_{\text{loss}}}} -\log p_{\theta,\bz}(\hat{s}_{k+1} | \hat{\bs}_{1:k}),
    \label{eq:loss}
\end{equation}
where $\Ical_{\text{loss}}$ is the set of token indices for computing loss.
The proposed latent tokens serve solely to assist the predictions of verbal tokens and thus we disregard the predictions of the latent tokens.
To this end, we minimize the loss of the predictions corresponding to verbal tokens, i.e.,
\begin{equation}
    \Ical_{\text{loss}} \triangleq \{ k\in\Rcal_{\bs}: \hat{s}_k \text{ is NOT a latent token} \}.
\end{equation}

\begin{remark}
    The proposed method is general and can be applied to any decoder-only Transformer model. Especially, we enable the parameter-efficient fine-tuning of latent tokens, which differs from the existing methods that require re-training the whole model from scratch~\cite{google2024think,pfau2024lets}.
    Besides, our method can use the existing infrastructure for training and inference, such as key-value (KV) caching and batch inference, with negligible additional infrastructure complexity.
\end{remark}

\subsection{Design Choices}\label{sec:design_choices}

\begin{figure*}[!t]
    \centering
    \includegraphics[width=0.92\linewidth]{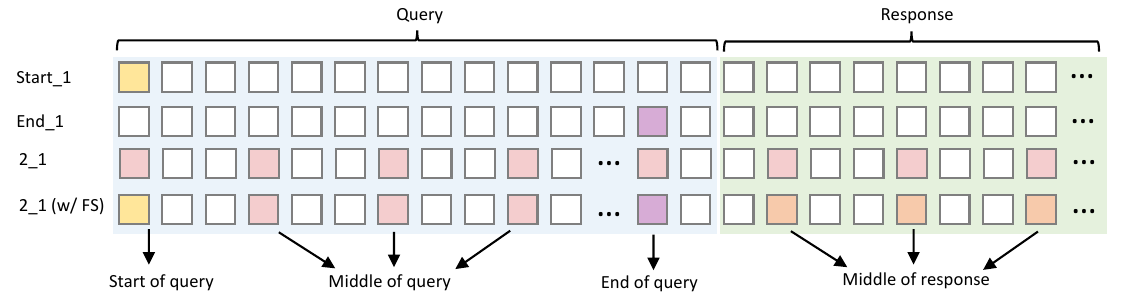}
    \caption{Several ways of inserting latent tokens into the sequence. The white squares denote verbal tokens, while those with different colors refer to different latent tokens. In particular, \start{1} and \eend{1} add one latent token at the start or end of the query, respectively; \freq{2}{1} inserts one latent token every two verbal tokens; \fs means function specialization. In \freq{2}{1} (w/ \fs), we show four groups of latent tokens with different functions for completeness; however, it is possible to select several groups to accommodate the need of the task.}
    \label{fig:latent_position}
    \vspace{-1em}
\end{figure*}

The latent tokens are incorporated into the sequence based on specific design choices. In principle, any number of latent tokens can be inserted at any position within the input sequence. Meanwhile, the number of latent tokens can be either fixed or variable across different input sequences. A straightforward strategy involves adding a group of $m$ latent tokens at a fixed frequency, e.g., once every $k$ verbal tokens, denoted by \freq{k}{m}.
Besides, latent tokens can be assigned based on specific input token markers, such as commas (,), periods (.), question marks (?), and others.
Some choices for inserting latent tokens are illustrated in Fig.~\ref{fig:latent_position}.

\paragraph{Function specialization.}
Furthermore, we introduce an advanced option that specializes latent tokens in different functions depending on their positions or followed verbal tokens. 
For example, we can use multiple groups of latent tokens at different positions of the sequence, e.g., the start of the query, the middle of the query, the end of the query, and the middle of the generated response.
Such a design of function specialization (denoted by \fs) ensures that the latent tokens inserted into multiple places will not suffer from conflicting learning objectives, thereby improving the model performance with a limited number of trainable parameters.

We summarize the relationships between the proposed method and several closely related works here:
\begin{itemize}[left=5pt]
    \item \emph{Prompt tuning}~\cite{prompt}: If only adding $m$ latent tokens at the start of the query sequence, denoted by \start{m}, the proposed method is equivalent to prompt tuning. In this case, the number of latent tokens can be regarded as the length of the soft prompt, which is used to guide the model in generating proper responses. However, our design is more general and can solve more complex tasks by inserting latent tokens at different positions, as will be demonstrated in Section~\ref{sec:synthetic} and Appendix~\ref{sec:benchmark}.\looseness=-1
    \item \emph{Pause token}~\cite{google2024think} and \emph{filler token}~\cite{pfau2024lets}: Pause token can be viewed as a variant of our method where $m$ latent tokens are appended at the end of the query, denoted by \eend{m}. In particular, some \pause tokens are expected as the CoT process before generating the final answer. Accordingly, the predictions of the input tokens are ignored until the last \pause token is seen.  
    Similarly, \cite{pfau2024lets} used filler tokens ($\dots$) to replace the CoT process in certain reasoning tasks, but found that it is hard to make the model learn the usage of filler tokens. 
    In comparison, our proposed latent tokens are broadly applicable for general tasks with only lightweight fine-tuning.
    In Appendix~\ref{sec:further_comparison}, we experimentally show that the proposed method achieves better performance than pause tokens due to the above designs.
\end{itemize}

It is worth noting that our proposed policies for inserting latent tokens are in line with the autoregressive property of LLM decoding,
i.e., whether a specific latent token should be inserted is determined by the context up to the current token.
In this work, we find that simple strategies, i.e., inserting latent tokens at some specific verbal token like a comma or even periodically, already secure satisfying performance in both synthetic tasks (Section~\ref{sec:synthetic}) and benchmark tasks (Appendix~\ref{sec:benchmark}).
In the following, we aim to investigate the types of tasks in which latent tokens offer advantages and the possible reasons behind this.\looseness=-1

\section{Exploring the Rationales Behind Latent Tokens}\label{sec:synthetic}

In this section, we explore the roles that latent tokens may play in the downstream tasks. We begin by proposing three hypotheses and then verify them through dedicated design of synthetic tasks. The detailed experimental setups and some supplementary experiments can be found in Appendix~\ref{sec:supple_synthetic}.

\subsection{Hypothesis 1: Self-Prompting for Long and Consistent Generation}

We hypothesize that latent tokens, when inserted repeatedly during autoregressive generation,
can serve the purpose of \emph{self-prompting}.
In other words, they continuously remind the Transformer model of the information that they learned from the training data,
which enables the model to generate long and consistent responses.

\paragraph{Setup.}
To verify this hypothesis, we design a task (Generation) where the model needs to generate a response sequence based on some rule for as long as possible.
This task requires generating equations based on a predefined rule learned from the training data. Specifically, we define an operation $@$ that takes four digits $a_1,a_2,b_1,b_2$ as inputs and the output results are $f_1(a_1,b_1) = |a_1 + b_1| \mod 9$ and $f_2(a_2,b_2) = |a_2 - b_2| \mod 9$. The equations are expressed as $a_1a_2@b_1b_2=f_1(a_1,b_1)f_2(a_2,b_2)$ followed by a comma, which in total takes six tokens with the tokenizer of \llamas model.
Then the equation proceeds by replacing $a_1a_2$ with $b_1b_2$ and $b_1b_2$ with $f_1(a_1,b_1)f_2(a_2,b_2)$.
The query in each sample contains the beginning token \bos and five equations.
Following this rule, the model is supposed to generate as many equations as it can within the maximal number of tokens in the response. However, this rule is not explicitly provided to the model, and the model needs to learn it from the training examples. 
The response of each training sample comprises several equations with lengths ranging from $1$ to $10$ without the ending token \eos.
One example is given as follows:

\begin{center}
\begin{minipage}{0.8\linewidth}
    \centering
    \fbox{%
      \parbox{\dimexpr\linewidth-2\fboxsep-2\fboxrule\relax}{%
      \query \bos 44@47=83,47@83=34,83@34=21,34@21=53,21@53=72,\\
      \response 53@72=31,72@31=11, 
    }%
    }
\end{minipage}
\end{center}

\begin{wrapfigure}{r}{0.48\textwidth}
    \includegraphics[width=\linewidth]{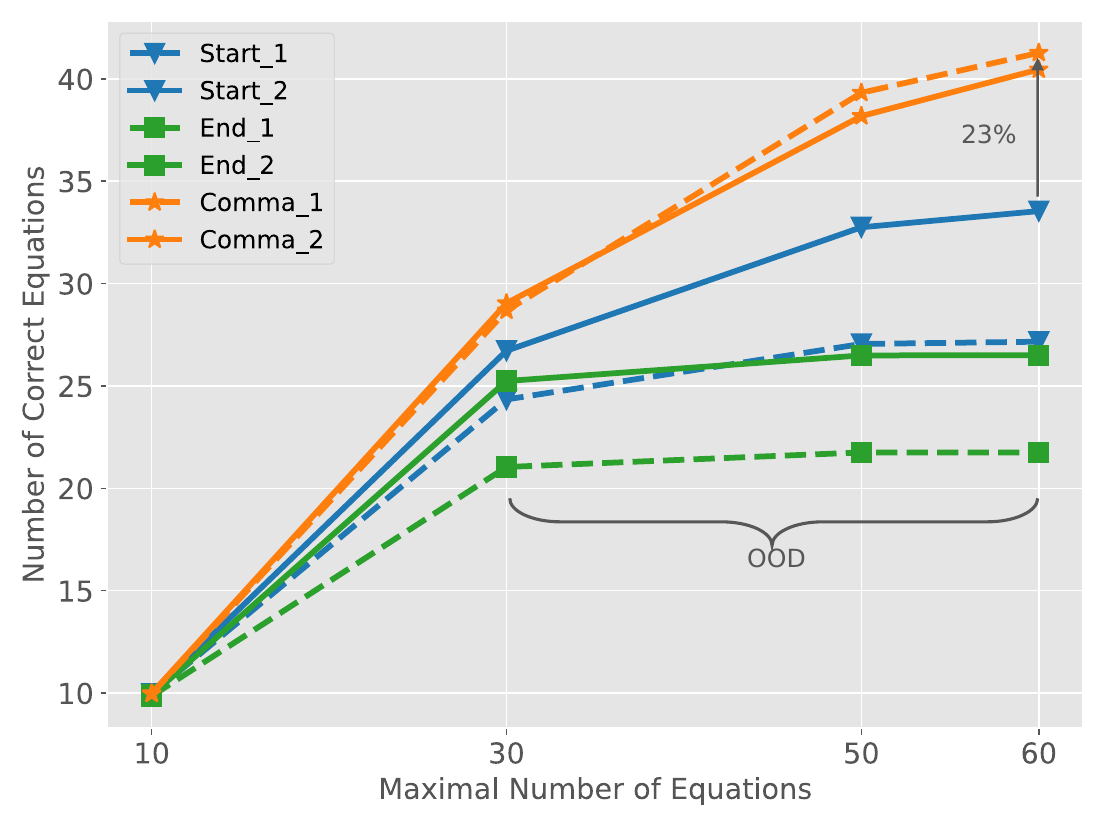}
    \caption{Numerical results for the Generation task, averaged over three random runs. The number $23\%$ represents a relative improvement over the best baseline.}
    \vspace{-0.5em}
    \label{fig:generation} 
\end{wrapfigure}

We implement several variants of the proposed method and observe the number of correct equations that can be generated by different methods in Fig.~\ref{fig:generation}.
The baselines \start{m} and \eend{m} refer to adding $m$ latent tokens at the start or end of the query, respectively; while the proposed method \comma{m} inserts $m$ latent tokens before each comma token.
We evaluate these methods in the in-distribution (ID) cases where the samples have the same distribution as the training data, as well as in the OOD cases where the Transformer needs to generate more equations than those of the training samples.\looseness=-1

\paragraph{Results.}

We observe from the results in Fig.~\ref{fig:generation} that \comma{1} and \comma{2} outperform all baselines under the same number of trainable parameters. 
In the OOD cases, the performance gap becomes wider as more equations are expected to be generated by the model.
In an extreme OOD setup, our method achieves a $23\%$ relative improvement over the best baseline.
One conceivable explanation is that by inserting latent tokens before each comma, the model receives contextual cues that effectively segment the equation into manageable components. This segmentation likely aids in maintaining coherence and structural integrity when generating long and consistent responses. In contrast, inserting latent tokens at the start or end of the query, as done in \start{m} and \eend{m}, provides less contextual information, limiting the model's ability to adapt to OOD scenarios. Consequently, the \comma{m} method enhances the model's generalization capabilities, enabling it to handle a broader range of equation numbers with greater accuracy.

\begin{figure*}[!t]
    \centering
    \begin{subfigure}{0.47\textwidth}
        \centering
        \includegraphics[width=\textwidth]{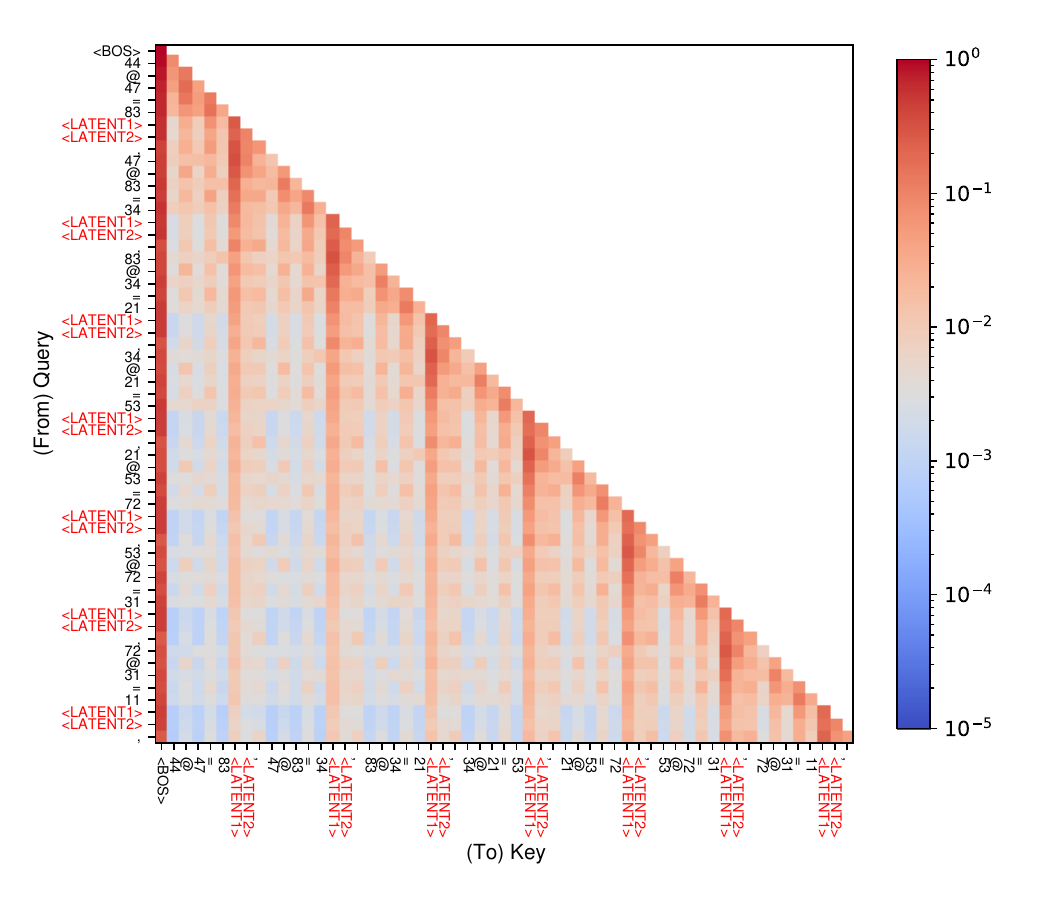}
        \caption{The first layer}
    \end{subfigure}
    \begin{subfigure}{0.47\textwidth}
        \centering
        \includegraphics[width=\textwidth]{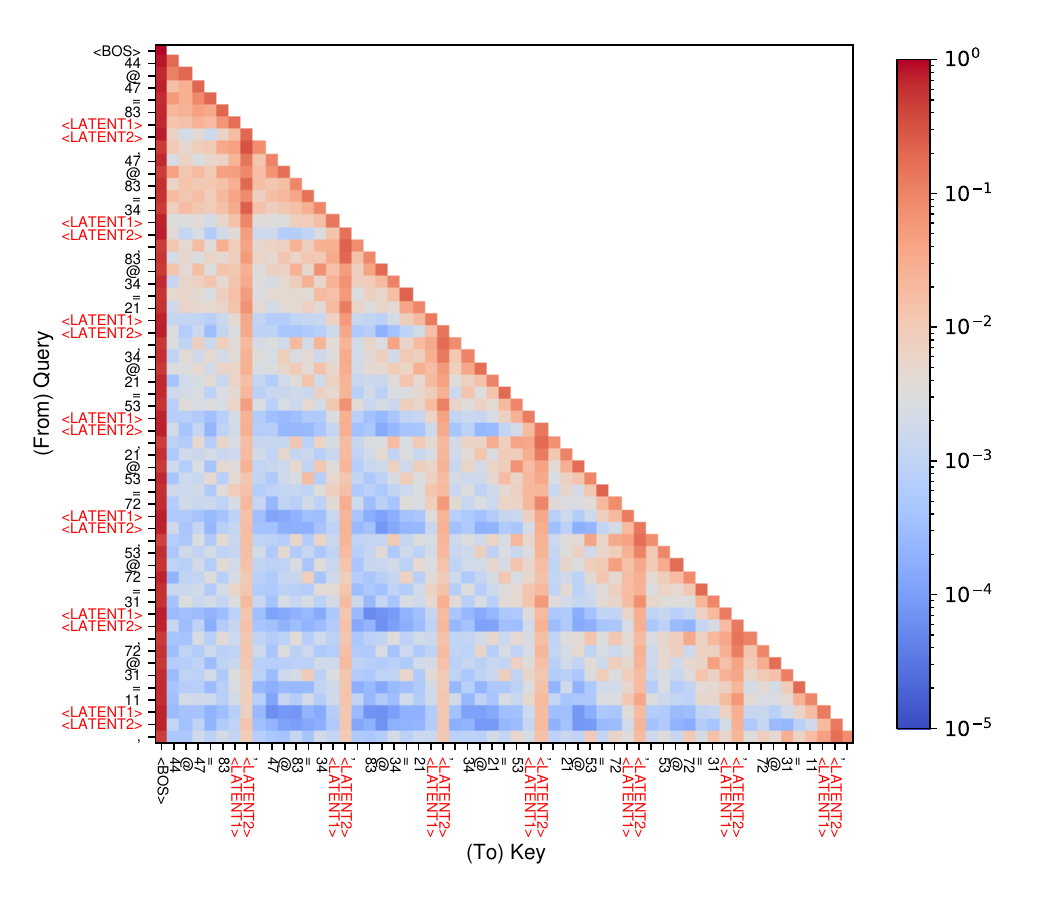}
        \caption{The last layer}
    \end{subfigure}
    \caption{Attention maps for the proposed \comma{2} method. Two latent tokens are denoted by {\color{red}\texttt{\textless LATENT1\textgreater}} and {\color{red}\texttt{\textless LATENT2\textgreater}}, respectively. For comparisons, the attention maps of other methods can be found in Appendix~\ref{sec:attention_maps}.}
    \label{fig:attention_map}
    \vspace{-0.5em}
\end{figure*}

To further investigate the role of the latent tokens, we visualize the attention maps of \comma{2} in Fig.~\ref{fig:attention_map}. 
We observe a pattern that repeats every eight tokens --- comprising six verbal tokens and two latent tokens --- which appears to become clearer in the last layer. Specifically, each group of latent tokens is heavily attended by a group of six subsequent tokens, implying its essential role in generating the following equation.\looseness=-1

\subsection{Hypothesis 2: Assisting Information Retrieval}

Our second hypothesis is that latent tokens inserted into the query can help the model retrieve relevant information from the input sequence.

\paragraph{Setup.}

To verify this hypothesis, we design a task (Summation) that requires the model to extract the relevant information from the input sequence and give the corresponding answer.
Each input sequence in this task involves a list of variables with specific integer values in the range of $[10,30)$. The end of the query is a question requesting the sum of two randomly selected variables from the variable list.
We note that computing the arithmetic sum of two variables is a simple task, but the challenge lies in extracting the correct information from the given sequence.
One example is given as:

\begin{center}
\begin{minipage}{0.9\linewidth}
\centering
\fbox{%
  \parbox{\dimexpr\linewidth-2\fboxsep-2\fboxrule\relax}{%
  \query \bos Var0=29, Var1=24, Var2=20, Var3=17, Var4=16, Var5=28, Var1+Var2=\\
  \response 24+20=44\eos
}%
}
\end{minipage}
\end{center}

We set the number of variables within the range $[5, 15)$ for each training sample, while evaluation can be conducted on test samples containing more variables, which can be regarded as OOD scenarios.

\paragraph{Results.}
\begin{wrapfigure}{r}{0.48\textwidth}
    \centering
    \includegraphics[width=\linewidth]{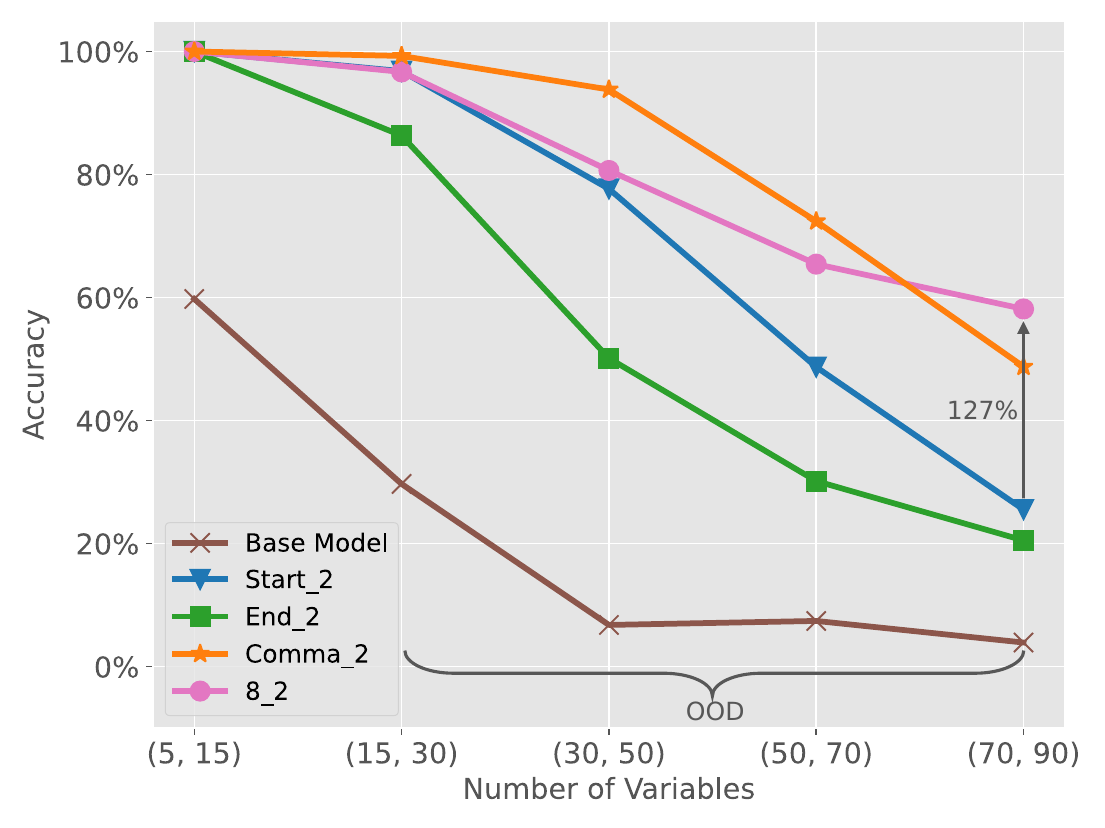}
    \caption{Numerical results for the Summation task, averaged over three random runs. The number $127\%$ represents a relative improvement over the best baseline.}
    \vspace{-0.5em}
    \label{fig:summation}
\end{wrapfigure} 

We vary the number of variables and show the performance of different methods, including several variants of the proposed method, in Fig.~\ref{fig:summation}.
The proposed method \comma{2} significantly outperforms the prompt tuning method (\start{2}), especially in OOD tests with a larger number of variables.
As the number of variables increases, the performance of all methods generally declines. However, it is noteworthy that the proposed method still achieves reasonable performance even when the number of variables is more than ten times greater than that used during training. 
Additionally, we evaluate the performance of periodically inserting the latent tokens, denoted by \freq{k}{m}, which means $m$ latent tokens for every $k$ verbal tokens. 
Interestingly, this periodic approach \freq{8}{2} also yields satisfactory results. In an extreme OOD setup, it achieves a $127\%$ relative improvement over the best baseline \start{2}.
Based on these findings, we conclude that latent tokens help in retrieving relevant information from input sequences. 
One possible explanation is that latent tokens act as ``anchors'' to locate the necessary information for answering questions, though a deeper investigation is needed to validate this hypothesis in the future.

\subsection{Hypothesis 3: Improving Instruction Adherence in Longer OOD Sequences}

We hypothesize that the usage of latent tokens during autoregressive generation can improve the LLM's ability to follow instructions, especially when required to generate extended sequences.

\paragraph{Setup.}
To validate this hypothesis, we design an instruction-following task (Repetition) where models are asked to repeat a given equation for a specified number of times.
Each sample is constructed as follows: we first generate a base equation of the form $a + b = c$, where $a$ and $b$ are randomly sampled integers between $1$ and $100$, and $c$ denotes the sum of $a$ and $b$. The query contains an explicit instruction ``Please repeat the following equations $n$ times:'', followed by the equation repeated $s_1$ times, where $s_1$ is randomly selected from $[1, 5]$. The model should continue the same equation exactly $s_2 = n \!-\! s_1$ more times, where $s_2$ ranges from $1$ to $10$. During training, only samples with $s_2 \in [1, 5]$ are used, while $s_2 > 5$ can be viewed as OOD evaluation.
One example is given as:\looseness=-1

\begin{center}
\begin{minipage}{0.88\linewidth}
    \centering
    \fbox{%
      \parbox{\dimexpr\linewidth-2\fboxsep-2\fboxrule\relax}{%
      \query \bos Please repeat the following equation 6 times: 48+19=67,48+19=67,\\
      \response 48+19=67,48+19=67,48+19=67,48+19=67,\eos
    }%
    }
\end{minipage}
\end{center}

Our preliminary studies find that the base model \llamas cannot finish this task without fine-tuning. 
Unlike the Generation task, the model is provided with the instruction, but needs to memorize the current generation status and strictly follow the repetition count in the initial instruction.
For this purpose, we propose the latent tokens with function specialization, namely \comma{1}~(w/~\fs), where a latent token is added at the start of the query, while the other latent token is regularly inserted before each comma.\looseness=-1

\paragraph{Results.}
\begin{wrapfigure}{r}{0.48\textwidth}
    \centering
    \includegraphics[width=\linewidth]{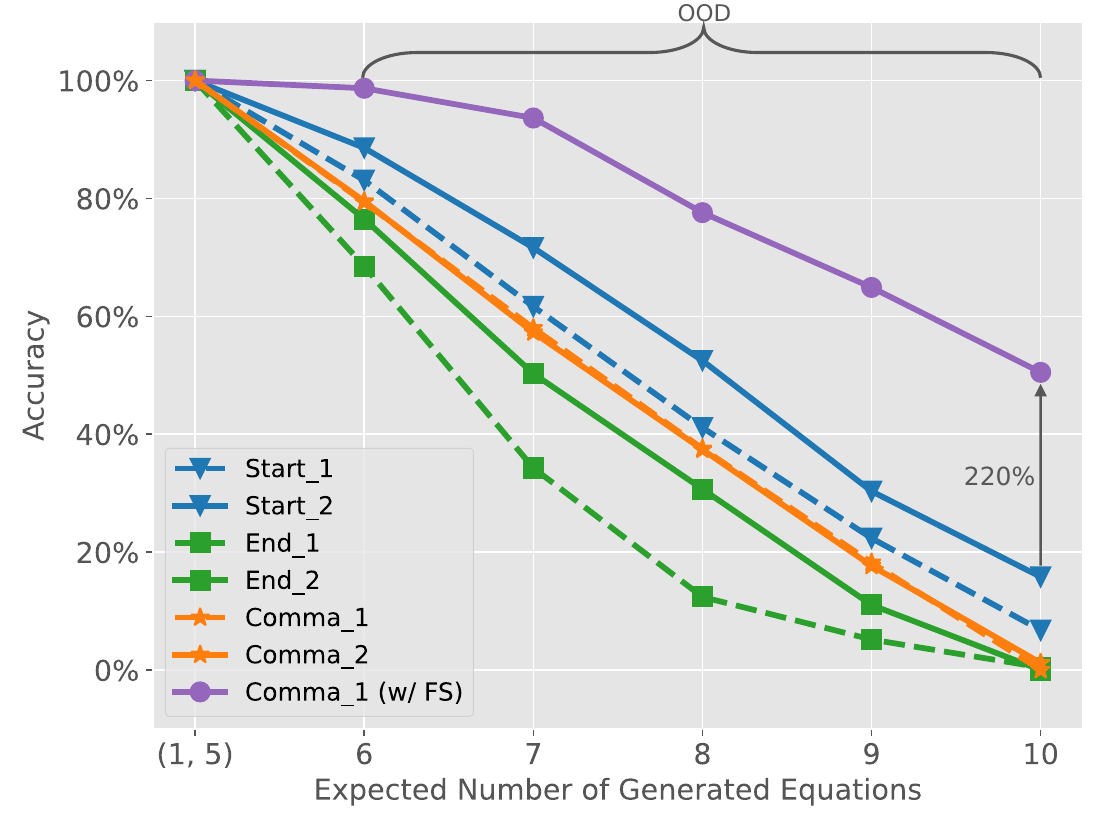}
    \caption{Numerical results for the Repetition task, averaged over three random runs. The number $220\%$ represents a relative improvement over the best baseline.}
    \vspace{-0.5em}
    \label{fig:repetition}
\end{wrapfigure}

From the experimental results in Fig.~\ref{fig:repetition}, we observe a severe performance drop in OOD scenarios compared to the 100\% accuracy in the ID setting. After checking the incorrect samples, we find that all failures are due to the wrong counting of equations, highlighting the difficulty of achieving the correct count when this count is unseen in training.
In contrast, the proposed method \comma{1} (w/ \fs) achieves significantly higher accuracy compared to all baselines, and in particular a $220\%$ relative improvement over the best baseline in an extreme OOD setup. 
This demonstrates the effectiveness of function specialization in adapting latent tokens for different functions. In particular, one latent token may memorize the instruction while others attend to previous latent tokens for accessing the current status. 
In Appendix~\ref{sec:attention_map_repetition}, we show some preliminary evidence that latent tokens may affect the ending of sequence generation by attending to the start latent token and instruction tokens.\looseness=-1

\subsection{Extended Comparisons}

In this subsection, we answer the following question: \textit{Does the performance gain of latent tokens simply result from the increased computation of extra tokens?}
To this end, we compare our method with the baselines that have the same average number of tokens as our method. Concretely, to match the average input sequence length used during training, we set the number of added tokens in the baselines to 23, 20, and 7 for the Generation, Summation, and Repetition tasks, respectively.

As shown in Table~\ref{table:prompt_synthetic}, our method significantly outperforms the baselines in the OOD scenarios, highlighting the generalization ability of the proposed method.
Moreover, we also observe that increasing the number of learnable tokens for the baseline methods does not necessarily lead to monotonic improvements in their performance, a phenomenon also reported in existing literature~\cite{lora}. 
In summary, the results indicate that the advantage of our method stems not only from increased computation but also from the more effective design of latent tokens.

\begin{table}[tbp]
\caption{Further comparisons with baselines. We report the number of equations for the Generation task and the accuracy for the Summation and Repetition tasks. All experiments are averaged over three random runs.}
\label{table:prompt_synthetic}
\centering
\resizebox{\textwidth}{!}{
\begin{tabular}{lcc|lcc|lcc}
\toprule
\multicolumn{3}{c|}{Generation} & \multicolumn{3}{c|}{Summation} & \multicolumn{3}{c}{Repetition} \\ 
\cmidrule(lr){1-3} \cmidrule(lr){4-6} \cmidrule(lr){7-9}
Method & 50 & 60 & Method & (30, 50) & (50, 70) & Method & 8 & 9 \\ 
\midrule
\rowcolor{cyan!15}
\comma{2} & \textbf{38.18} & \textbf{40.46} & \comma{2} & \textbf{93.82\%} & \textbf{72.40\%} & \comma{1} (w/ \fs) & \textbf{77.59\%} & \textbf{64.90\%} \\
\start{23} & 23.17 & 23.45 & \start{20} & \underline{53.32\%} & \underline{27.86\%} & \start{7} & \underline{48.47\%} & \underline{29.95\%} \\
\eend{23} & \underline{27.49} & \underline{27.53} & \eend{20} & 53.12\% & 22.92\% & \eend{7} & 29.80\% & 12.47\% \\
\bottomrule
\end{tabular}}
\end{table}

\section{Related Works}

\paragraph{Learnable special tokens.}
Incorporating some learnable tokens into the Transformers has been widely explored for different purposes, including memorizing previous sequence~\cite{bulatov2022memory}, offering adaptive computation~\cite{adatape}, compressing the prompts~\cite{mu2024gist}, and mitigating certain artifacts in the feature maps of vision Transformers \cite{regitster}.
To improve the reasoning ability, \cite{google2024think} proposed to insert some \pause tokens within or at the end of the query.
\cite{pfau2024lets} use filler tokens ($\dots$) to replace the CoT process while improving the performance compared with non-CoT cases in some designed tasks.
However, these works find it hard to achieve the desired performance by fine-tuning these special tokens directly~\cite{vennam2024rethinking}, which motivates us to design a more effective approach to introducing latent computation.
Besides, \cite{wang2023guiding} trains the LLM to generate a planning token at the start of each reasoning step, which guides the model for better reasoning. 
These studies mainly focus on tasks with explicit reasoning steps, limiting their generalization to other tasks.
Furthermore, prompt tuning~\cite{prompt} can also be viewed as prior works of prepending or appending special tokens to the input query, which is a special case of our proposed method.

\paragraph{Broader Related Works.}
A growing body of research demonstrates that integrating intermediate computation steps can enhance the capabilities of LLMs on multiple downstream applications
~\cite{snell2024scaling,welleck2024from,ji2025testtimecomputingsystem1thinking,zeng2024scalingsearchlearningroadmap}.
For example, many studies, including scratchpad~\cite{scratchpad}, CoT prompting~\cite{cot,william2024expressive}, and so on~\cite{tot,quiet-star}, prompt LLMs to generate some intermediate computation steps explicitly to improve the reasoning ability of the model.
In this case, the LLMs are restricted to reasoning in ``natural language space'' and thus may achieve limited performance in some cases ~\cite{coconut}.
Another line of research proposes to use the hidden computation to guide the model output, while the results of hidden computation may not have interpretable meanings in natural language~\cite{coconut,liu2024deliberation}.
These methods effectively augment the reasoning abilities but typically require heavy training of the whole Transformer model, making them hard to adapt to different downstream tasks.

\section{Conclusions and Discussions}\label{sec:conclusion}

In this work, we introduce a unified approach to integrating latent tokens into LLMs and thereby provide additional latent computation, with a dedicated design for various aspects such as positional encoding and loss function.
The proposed method is generally applicable to any decoder-only Transformer and compatible with existing infrastructures for training and inference. 
Thanks to the lightweight nature of the latent tokens, batch inference with multiple sets of learned latent tokens can be easily applied for serving different downstream tasks or users simultaneously.
Moreover, we delve into the underlying mechanisms of latent tokens in LLM generation by proposing and validating three possible hypotheses through synthetic tasks. Importantly, these results show the OOD generalization ability of latent tokens in various tasks. 

There are some works requiring future investigations. First, the design space of latent tokens involves tuning some hyper-parameters, but our experiments have not exhaustively covered all possible configurations. 
Besides, this work primarily focuses on simple policies of adding latent tokens. It would be intriguing to develop more adaptive and advanced strategies to fully unlock their potential.
Additionally, despite that we have explored some possible working mechanisms of latent tokens via synthetic tasks, a complete understanding of their underlying rationales is left to future work.
Lastly, we adopt supervised fine-tuning as the training scheme, but future work may extend the optimization process to other settings, such as reinforcement fine-tuning.

\clearpage

\bibliographystyle{plain}
\bibliography{refs}

\begin{thebibliography}{10}

\bibitem{pitfalls}
Gregor Bachmann and Vaishnavh Nagarajan.
\newblock The pitfalls of next-token prediction.
\newblock In {\em The Twelfth International Conference on Learning Representations (ICLR)}, Vienna, Austria, May 2024.

\bibitem{wikisplit}
Jan~A Botha, Manaal Faruqui, John Alex, Jason Baldridge, and Dipanjan Das.
\newblock Learning to split and rephrase from wikipedia edit history.
\newblock In {\em Proceedings of the 2018 Conference on Empirical Methods in Natural Language Processing}, pages 732--737, Brussels, Belgium, Oct.-Nov. 2018.

\bibitem{BrownMRSKDNSSAA20}
Tom~B. Brown, Benjamin Mann, Nick Ryder, Melanie Subbiah, Jared Kaplan, Prafulla Dhariwal, Arvind Neelakantan, Pranav Shyam, Girish Sastry, Amanda Askell, Sandhini Agarwal, Ariel Herbert{-}Voss, Gretchen Krueger, Tom Henighan, Rewon Child, Aditya Ramesh, Daniel~M. Ziegler, Jeffrey Wu, Clemens Winter, Christopher Hesse, Mark Chen, Eric Sigler, Mateusz Litwin, Scott Gray, Benjamin Chess, Jack Clark, Christopher Berner, Sam McCandlish, Alec Radford, Ilya Sutskever, and Dario Amodei.
\newblock Language models are few-shot learners.
\newblock In {\em Advances in Neural Information Processing Systems (NeurIPS)}, Virtual Event, Dec. 2020.

\bibitem{bulatov2022memory}
Aydar Bulatov, Yuri Kuratov, and Mikhail Burtsev.
\newblock Recurrent memory transformer.
\newblock In {\em Advances in Neural Information Processing Systems (NeurIPS)}, New Orleans, LA, {USA}, Nov. 2022.

\bibitem{gsm8k}
Karl Cobbe, Vineet Kosaraju, Mohammad Bavarian, Mark Chen, Heewoo Jun, Lukasz Kaiser, Matthias Plappert, Jerry Tworek, Jacob Hilton, Reiichiro Nakano, Christopher Hesse, and John Schulman.
\newblock Training verifiers to solve math word problems.
\newblock {\em arXiv preprint arXiv:2110.14168}, 2021.

\bibitem{regitster}
Timoth{\'{e}}e Darcet, Maxime Oquab, Julien Mairal, and Piotr Bojanowski.
\newblock Vision transformers need registers.
\newblock In {\em The Twelfth International Conference on Learning Representations (ICLR)}, Vienna, Austria, May 2024.

\bibitem{faith}
Nouha Dziri, Ximing Lu, Melanie Sclar, Xiang~Lorraine Li, Liwei Jiang, Bill~Yuchen Lin, Sean Welleck, Peter West, Chandra Bhagavatula, Ronan~Le Bras, Jena~D. Hwang, Soumya Sanyal, Xiang Ren, Allyson Ettinger, Za{\"{\i}}d Harchaoui, and Yejin Choi.
\newblock Faith and fate: Limits of transformers on compositionality.
\newblock In {\em Advances in Neural Information Processing Systems (NeurIPS)}, New Orleans, LA, {USA}, Nov. 2023.

\bibitem{google2024think}
Sachin Goyal, Ziwei Ji, Ankit~Singh Rawat, Aditya~Krishna Menon, Sanjiv Kumar, and Vaishnavh Nagarajan.
\newblock {Think before you speak: Training Language Models With Pause Tokens}.
\newblock In {\em The Twelfth International Conference on Learning Representations (ICLR)}, Vienna, Austria, May 2024.

\bibitem{coconut}
Shibo Hao, Sainbayar Sukhbaatar, DiJia Su, Xian Li, Zhiting Hu, Jason Weston, and Yuandong Tian.
\newblock Training large language models to reason in a continuous latent space.
\newblock {\em arXiv preprint arXiv:2412.06769}, 2024.

\bibitem{lora}
Edward~J. Hu, Yelong Shen, Phillip Wallis, Zeyuan Allen{-}Zhu, Yuanzhi Li, Shean Wang, Lu~Wang, and Weizhu Chen.
\newblock {LoRA}: {Low}-rank adaptation of large language models.
\newblock In {\em The Tenth International Conference on Learning Representations (ICLR)}, Virtual Event, Apr. 2022.

\bibitem{ji2025testtimecomputingsystem1thinking}
Yixin Ji, Juntao Li, Hai Ye, Kaixin Wu, Jia Xu, Linjian Mo, and Min Zhang.
\newblock Test-time computing: from system-1 thinking to system-2 thinking.
\newblock {\em arXiv preprint arXiv:2501.02497}, 2025.

\bibitem{ke2021rethinking}
Guolin Ke, Di~He, and Tie{-}Yan Liu.
\newblock Rethinking positional encoding in language pre-training.
\newblock In {\em 9th International Conference on Learning Representations (ICLR)}, Vienna, Austria, May 2021.

\bibitem{narrativeqa}
Tom{\'{a}}s Kocisk{\'{y}}, Jonathan Schwarz, Phil Blunsom, Chris Dyer, Karl~Moritz Hermann, G{\'{a}}bor Melis, and Edward Grefenstette.
\newblock The {NarrativeQA} reading comprehension challenge.
\newblock {\em Transactions of the Association for Computational Linguistics}, 6:317--328, 2018.

\bibitem{prompt}
Brian Lester, Rami Al{-}Rfou, and Noah Constant.
\newblock The power of scale for parameter-efficient prompt tuning.
\newblock In {\em Proceedings of the 2021 Conference on Empirical Methods in Natural Language Processing (EMNLP)}, pages 3045--3059, Punta Cana, Dominican Republic, Nov. 2021.

\bibitem{liu2024deliberation}
Luyang Liu, Jonas Pfeiffer, Jiaxing Wu, Jun Xie, and Arthur Szlam.
\newblock Deliberation in latent space via differentiable cache augmentation.
\newblock {\em arXiv preprint arXiv:2412.17747}, 2024.

\bibitem{meta2024llama3herdmodels}
{Llama Team, AI @ Meta}.
\newblock The {Llama} 3 herd of models.
\newblock {\em arXiv preprint arXiv:2407.21783}, 2024.

\bibitem{william2024expressive}
William Merrill and Ashish Sabharwal.
\newblock The expressive power of transformers with chain of thought.
\newblock In {\em The Twelfth International Conference on Learning Representations (ICLR)}, Vienna, Austria, May 2024.

\bibitem{mu2024gist}
Jesse Mu, Xiang Li, and Noah Goodman.
\newblock Learning to compress prompts with gist tokens.
\newblock In {\em Advances in Neural Information Processing Systems (NeurIPS)}, Vancouver, Canada, 2024.

\bibitem{NowakSBC24}
Franz Nowak, Anej Svete, Alexandra Butoi, and Ryan Cotterell.
\newblock On the representational capacity of neural language models with chain-of-thought reasoning.
\newblock In {\em Proceedings of the 62nd Annual Meeting of the Association for Computational Linguistics (ACL)}, pages 12510--12548, Bangkok, Thailand, Aug. 2024.

\bibitem{scratchpad}
Maxwell Nye, Anders~Johan Andreassen, Guy Gur-Ari, Henryk Michalewski, Jacob Austin, David Bieber, David Dohan, Aitor Lewkowycz, Maarten Bosma, David Luan, Charles Sutton, and Augustus Odena.
\newblock Show your work: Scratchpads for intermediate computation with language models.
\newblock {\em arXiv preprint arXiv:2112.00114}, 2021.

\bibitem{openai2024gpt4}
OpenAI.
\newblock {GPT-4} technical report.
\newblock {\em arXiv preprint arXiv:2303.08774}, 2024.

\bibitem{pfau2024lets}
Jacob Pfau, William Merrill, and Samuel~R. Bowman.
\newblock Let's think dot by dot: Hidden computation in transformer language models.
\newblock In {\em First Conference on Language Modeling (COLM)}, Philadelphia, PA, USA, Oct. 2024.

\bibitem{snell2024scaling}
Charlie Snell, Jaehoon Lee, Kelvin Xu, and Aviral Kumar.
\newblock Scaling llm test-time compute optimally can be more effective than scaling model parameters.
\newblock {\em arXiv preprint arXiv:2408.03314}, 2024.

\bibitem{su2024roformer}
Jianlin Su, Murtadha Ahmed, Yu~Lu, Shengfeng Pan, Wen Bo, and Yunfeng Liu.
\newblock Roformer: Enhanced transformer with rotary position embedding.
\newblock {\em Neurocomputing}, 568:127063, 2024.

\bibitem{vaswani2017transformer}
Ashish Vaswani, Noam Shazeer, Niki Parmar, Jakob Uszkoreit, Llion Jones, Aidan~N Gomez, {\L}ukasz Kaiser, and Illia Polosukhin.
\newblock Attention is all you need.
\newblock In {\em Advances in Neural Information Processing Systems (NIPS)}, Long Beach, CA, USA, Dec. 2017.

\bibitem{vennam2024rethinking}
Sreeram Vennam, David Valente, David Herel, and Ponnurangam Kumaraguru.
\newblock Rethinking thinking tokens: Understanding why they underperform in practice.
\newblock {\em arXiv preprint arXiv:2411.11371}, 2024.

\bibitem{wang2023guiding}
Xinyi Wang, Lucas Caccia, Oleksiy Ostapenko, Xingdi Yuan, William~Yang Wang, and Alessandro Sordoni.
\newblock Guiding language model reasoning with planning tokens.
\newblock In {\em First Conference on Language Modeling (COLM)}, Philadelphia, PA, USA, Oct. 2024.

\bibitem{cot}
Jason Wei, Xuezhi Wang, Dale Schuurmans, Maarten Bosma, Brian Ichter, Fei Xia, Ed~H. Chi, Quoc~V. Le, and Denny Zhou.
\newblock Chain-of-thought prompting elicits reasoning in large language models.
\newblock In {\em Advances in Neural Information Processing Systems (NeurIPS)}, New Orleans, LA, {USA}, Nov. 2022.

\bibitem{welleck2024from}
Sean Welleck, Amanda Bertsch, Matthew Finlayson, Hailey Schoelkopf, Alex Xie, Graham Neubig, Ilia Kulikov, and Zaid Harchaoui.
\newblock From decoding to meta-generation: Inference-time algorithms for large language models.
\newblock {\em Transactions on Machine Learning Research}, 2024.

\bibitem{adatape}
Fuzhao Xue, Valerii Likhosherstov, Anurag Arnab, Neil Houlsby, Mostafa Dehghani, and Yang You.
\newblock Adaptive computation with elastic input sequence.
\newblock In {\em International Conference on Machine Learning (ICML)}, pages 38971--38988, Honolulu, HI, {USA}, July 2023.

\bibitem{yan2024understanding}
Jianhao Yan, Jin Xu, Chiyu Song, Chenming Wu, Yafu Li, and Yue Zhang.
\newblock Understanding in-context learning from repetitions.
\newblock In {\em The Twelfth International Conference on Learning Representations (ICLR)}, Vienna, Austria, May 2024.

\bibitem{tot}
Shunyu Yao, Dian Yu, Jeffrey Zhao, Izhak Shafran, Tom Griffiths, Yuan Cao, and Karthik Narasimhan.
\newblock Tree of thoughts: Deliberate problem solving with large language models.
\newblock In {\em Advances in Neural Information Processing Systems (NeurIPS)}, New Orleans, LA, {USA}, Nov. 2023.

\bibitem{quiet-star}
Eric Zelikman, Georges Harik, Yijia Shao, Varuna Jayasiri, Nick Haber, and Noah~D Goodman.
\newblock Quiet-star: Language models can teach themselves to think before speaking.
\newblock {\em arXiv preprint arXiv:2403.09629}, 2024.

\bibitem{zeng2024scalingsearchlearningroadmap}
Zhiyuan Zeng, Qinyuan Cheng, Zhangyue Yin, Bo~Wang, Shimin Li, Yunhua Zhou, Qipeng Guo, Xuanjing Huang, and Xipeng Qiu.
\newblock Scaling of search and learning: A roadmap to reproduce o1 from reinforcement learning perspective.
\newblock {\em arXiv preprint arXiv:2412.14135}, 2024.

\bibitem{zhao2024length}
Liang Zhao, Xiachong Feng, Xiaocheng Feng, Weihong Zhong, Dongliang Xu, Qing Yang, Hongtao Liu, Bing Qin, and Ting Liu.
\newblock Length extrapolation of transformers: {A} survey from the perspective of positional encoding.
\newblock In {\em Findings of the Association for Computational Linguistics (EMNLP)}, pages 9959--9977, Miami, FL, USA, Nov. 2024.

\end{thebibliography}

\clearpage

\appendix

\section{Design Details}\label{sec:design_details}

\subsection{Analysis on Positional Encoding for Latent Tokens}\label{sec:position_encoding}

In this subsection, we compare the positional encoding schemes for a standard Transformer, naive positional encoding with directly inserting latent tokens, and our proposed positional encoding. Their differences are illustrated in Fig.~\ref{fig:position_IDs}.

\paragraph{Preliminaries and notations.}

Suppose that we are given a context consisting of $t$ standard verbal tokens, with
hidden states (at a certain attention layer) $\bx_1, \bx_2, \dots, \bx_t$ and position IDs $1, 2, \dots, t$. The self-attention module incorporates position information into the input hidden states and transforms them into query, key, and value representations, which are respectively given by:
\begin{equation}
    \bq_t = f_q(\bx_t, t),\quad 
    \bk_i = f_k(\bx_i, i),\quad 
    \bv_i = f_v(\bx_i, i),
\end{equation}
where $f_q$, $f_k$, and $f_v$ are the query, key, and value functions, respectively. The query $\bq_t$ and key $\bk_i$ are then used to compute the attention weights, defined as $\alpha_{t, i}$, while the output is computed as the weighted sum over the values $\bv_i$:
\begin{align}
    \alpha_{t, i} &= \frac{\exp \left(\frac{\bq_t^\trans \bk_i}{\sqrt{d}} \right)}{\sum_{j=1}^n \exp \left(\frac{\bq_t^\trans \bk_j}{\sqrt{d}} \right)},\\
    \bo_t &= \sum_{i=1}^n \alpha_{t, i} \bv_i,
\end{align}
where $d$ is the dimension of the query and key vectors. The output of the self-attention module is then passed through a feed-forward neural network to generate the output of this Transformer block. Such a process is repeated for multiple layers of the Transformer, and the final output is fed into the softmax layer to predict the output tokens.

For notational convenience, we leverage the notation $g$ in the following:
$g(\bx_m, \bx_n, m, n) = \frac{1}{\sqrt{d}} \langle f_q(\bx_m, m), f_k(\bx_n, n) \rangle$.

\paragraph{Standard Transformer.}

Consider the forward pass for the $(t+1)$-th token at a certain self-attention layer, 
with input hidden state $\bx_{t+1}$.
The corresponding output is
\begin{align}
    \label{eq:output_standard}
    \bo_{t+1} &= \sum_{i \in [t+1]} \alpha_i f_v(\bx_i, i) 
    = \sum_{i \in [t]} \alpha_i f_v(\bx_i, i) + \alpha_{t+1} f_v(\bx_{t+1}, t+1), 
\end{align}
where
\begin{subequations}
    \label{eq:alpha_standard}
    \begin{align}
        \balpha = \softmax\bigg(\concat\Big(
            &\big[ g(\bx_{t+1}, \bx_i, t+1, i) \big]_{i \in [t]},  \\
            &\big[ g(\bx_{t+1}, \bx_{t+1}, t+1, t+1) \big]
            \Big)\bigg).    
    \end{align}
\end{subequations}

\paragraph{Naive positional encoding for latent tokens.}

Suppose that $m$ latent tokens with hidden states $\bz_1, \dots, \bz_m$ are inserted before the forward pass for $\bx_{t+1}$.
With naive positional encoding,
the tokens in the context $\bx_1, \dots, \bx_t, \bz_1, \dots, \bz_m$ 
have position IDs $1, \dots, t, t+1, \dots, t+m$.
In this case, the output will be 
\begin{align*}
    \bo_{t+1} &= \sum_{i \in [t]} \alpha_i f_v(\bx_i, i) 
    + \sum_{i \in [m]} \alpha_{t+i} f_v(\bz_i, t+i)
    + \alpha_{t+m+1} f_v(\bx_{t+1}, t+m+1), 
\end{align*}
where
\begin{subequations}
    \label{eq:alpha_naive_pos_id}
    \begin{align}
        \balpha = \softmax\bigg(\concat\Big(
            &\big[ g(\bx_{t+1}, \bx_i, t+m+1, i) \big]_{i \in [t]},  \\
            &\big[ g(\bx_{t+1}, \bz_i, t+m+1, t+i) \big]_{i \in [m]}, \\
            &\big[ g(\bx_{t+1}, \bx_{t+1}, t+m+1, t+m+1) \big]
            \Big)\bigg).   
    \end{align}
\end{subequations}

For a large value of $m$,
this is a significant drift from the original pre-trained Transformer,
which can make it challenging to train the latent tokens alone while freezing all modules of the pre-trained model.

\paragraph{Our proposed positional encoding.}

We propose to freeze the position IDs when latent tokens are inserted.
With this approach, the tokens in the context $\bx_1, \dots, \bx_t, \bz_1, \dots, \bz_m$ 
have position IDs $1, \dots, t, t+1, \dots, t+1$,
i.e., the last $m$ position IDs are all $t+1$, same as that for $\bx_{t+1}$.

In this case, the output will be
\begin{align}
    \label{eq:output_proposed_pos_id}
    \bo_{t+1} = \sum_{i \in [t]} \alpha_i f_v(\bx_i, i) 
    + \sum_{i \in [m]} \alpha_{t+i} f_v(\bz_i, t+1)
    + \alpha_{t+m+1} f_v(\bx_{t+1}, t+1),  
\end{align}
where
\begin{subequations}
    \label{eq:alpha_proposed_pos_id}
    \begin{align}
        \balpha = \softmax\bigg(\concat\Big(
            &\big[ g(\bx_{t+1}, \bx_i, t+1, i) \big]_{i \in [t]},  \\
            &\big[ g(\bx_{t+1}, \bz_i, t+1, t+1) \big]_{i \in [m]}, \\
            &\big[ g(\bx_{t+1}, \bx_{t+1}, t+1, t+1) \big] 
            \Big)\bigg).   
    \end{align}
\end{subequations}
Comparing Eq.~\eqref{eq:output_proposed_pos_id} with Eq.~\eqref{eq:output_standard}, 
as well as Eq.~\eqref{eq:alpha_proposed_pos_id} with Eq.~\eqref{eq:alpha_standard},
we note that the only difference of the proposed positional encoding from the standard Transformer is in the middle term corresponding to the newly added latent tokens.
With such a design, the Transformer augmented with randomly initialized latent tokens remains close to the pre-trained standard Transformer,
which facilitates stable training of the latent tokens while the original modules are frozen.
Our proposed positional encoding achieves a minimal disturbance of position IDs from that of standard transformers, allowing them to be integrated seamlessly, whereas the naive method 
disrupts the position distribution of subsequent verbal tokens after a group of latent tokens.

\subsection{Prepending or Appending Latent Tokens}\label{sec:append_or_prepend}

\begin{figure}[htbp]
    \centering
    \includegraphics[width=\textwidth]{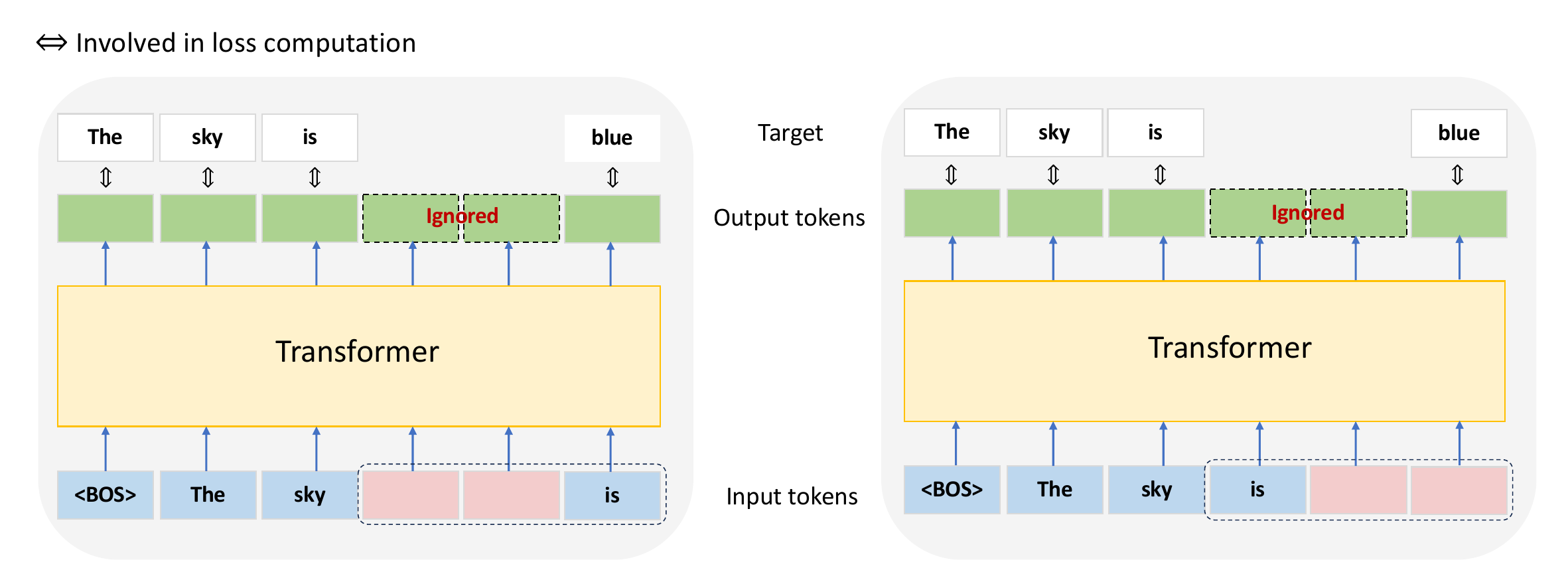}
    \caption{Comparisons between prepending latent tokens (\prepend) and appending latent tokens (\append).}
    \label{fig:compare-loss}
\end{figure}

In this subsection, we briefly discuss another choice of inserting the latent tokens into the sequence. Concretely, we may append a group of latent tokens after a verbal token $s_t$ and assign the position ID $t$ to these latent tokens.
Then, in both training and inference processes, the predictions of the Transformer are ignored until the last latent token is seen~\cite{google2024think}, as illustrated in Fig.~\ref{fig:compare-loss}. We refer to such a choice as ``appending latent tokens'' (denoted by \append) in the following, and the default setup as ``prepending latent tokens'' (denoted by \prepend).
The experiments in Appendix~\ref{sec:ablation} provide a comparison between these two designs.

\clearpage

\section{Supplementary Materials for Synthetic Tasks}\label{sec:supple_synthetic}
\subsection{Experimental Setup}\label{sec:exp_details_synthetic}

We implement all methods with Pytorch\footnote{\url{https://github.com/pytorch/pytorch}} and Transformers\footnote{\url{https://github.com/huggingface/Transformers}} packages and run experiments on Nvidia A100 and A800 GPUs. All results are averaged over three random seeds.

We use the pre-trained \llamas as the base model on the synthetic tasks.
The values of hyperparameters are summarized in Table~\ref{table:hyperparameters_synthetic}.

\begin{table}[h]
    \caption{Experimental setup for synthetic tasks.}
    \label{table:hyperparameters_synthetic}
    \centering
    \begin{tabular}{l|ccc}
    \toprule
    Dataset                          & Generation & Summation & Repetition \\ \midrule
    \multicolumn{4}{c}{\textbf{Training}} \\ \midrule
    Optimizer                        & \multicolumn{3}{c}{AdamW}     \\ 
    Weight Decay                     &  \multicolumn{3}{c}{1e-4}   \\ 
    \# of Epochs                     &  \multicolumn{3}{c}{10}          \\ 
    Warmup Ratio                     &  \multicolumn{3}{c}{0.01}          \\ 
    Learning Rate Schedule           &  \multicolumn{3}{c}{Cosine}        \\ 
    Learning Rate                    &  \multicolumn{3}{c}{5e-2}       \\
    Minimal Learning Rate            &  \multicolumn{3}{c}{5e-5}       \\
    Batch Size                       & \multicolumn{3}{c}{16}        \\ 
    \# of Training Samples           &\multicolumn{3}{c}{4096}          \\ 
    \# of Validation Samples         & \multicolumn{3}{c}{1024}     \\\midrule
    \multicolumn{4}{c}{\textbf{Inference}} \\ \midrule
    \# of Test Samples               & \multicolumn{3}{c}{1024}         \\
    Maximal \# of New Tokens         & 30/180/300/360 & 10 & 200      \\ 
    \bottomrule
    \end{tabular}
\end{table}

\subsection{Supplementary Attention Maps for the Generation Task}\label{sec:attention_maps}
\begin{figure*}[htbp]
    \centering
    \begin{subfigure}{0.48\textwidth}
        \centering
        \includegraphics[width=\textwidth]{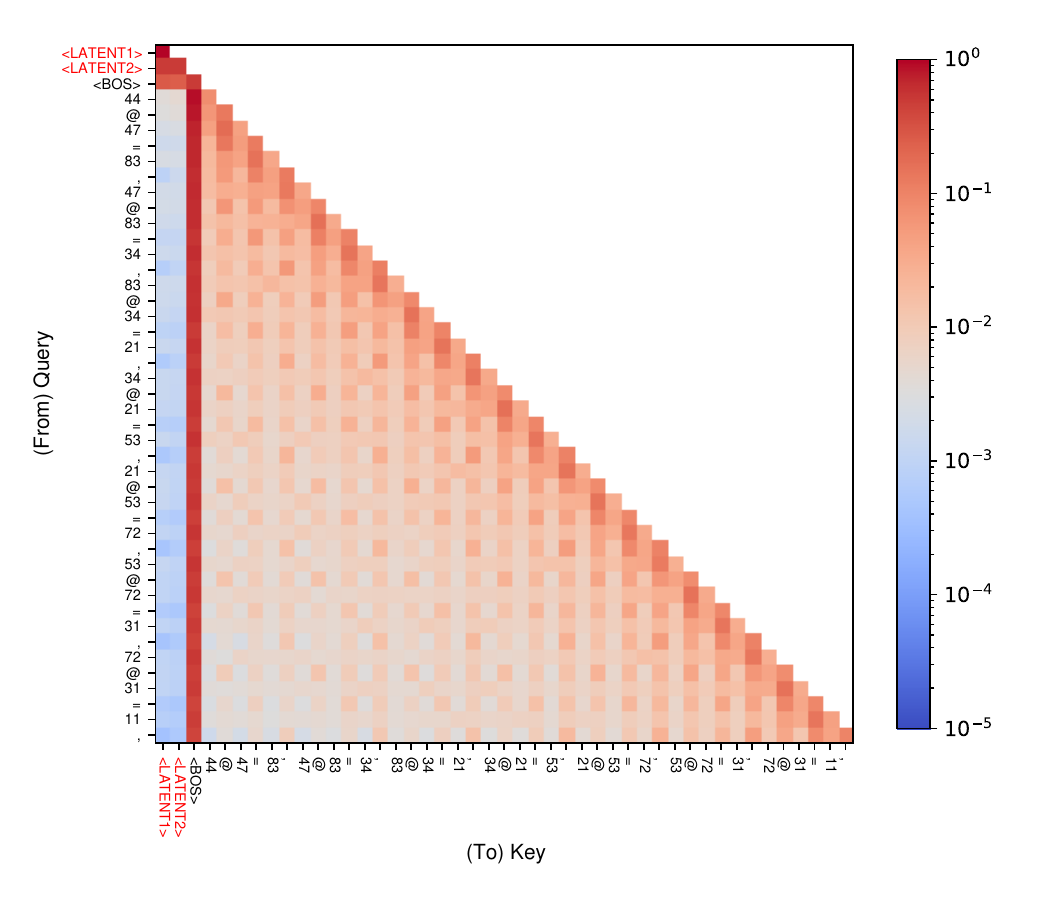}
        \caption{The first layer}
    \end{subfigure}
    \begin{subfigure}{0.48\textwidth}
        \centering
        \includegraphics[width=\textwidth]{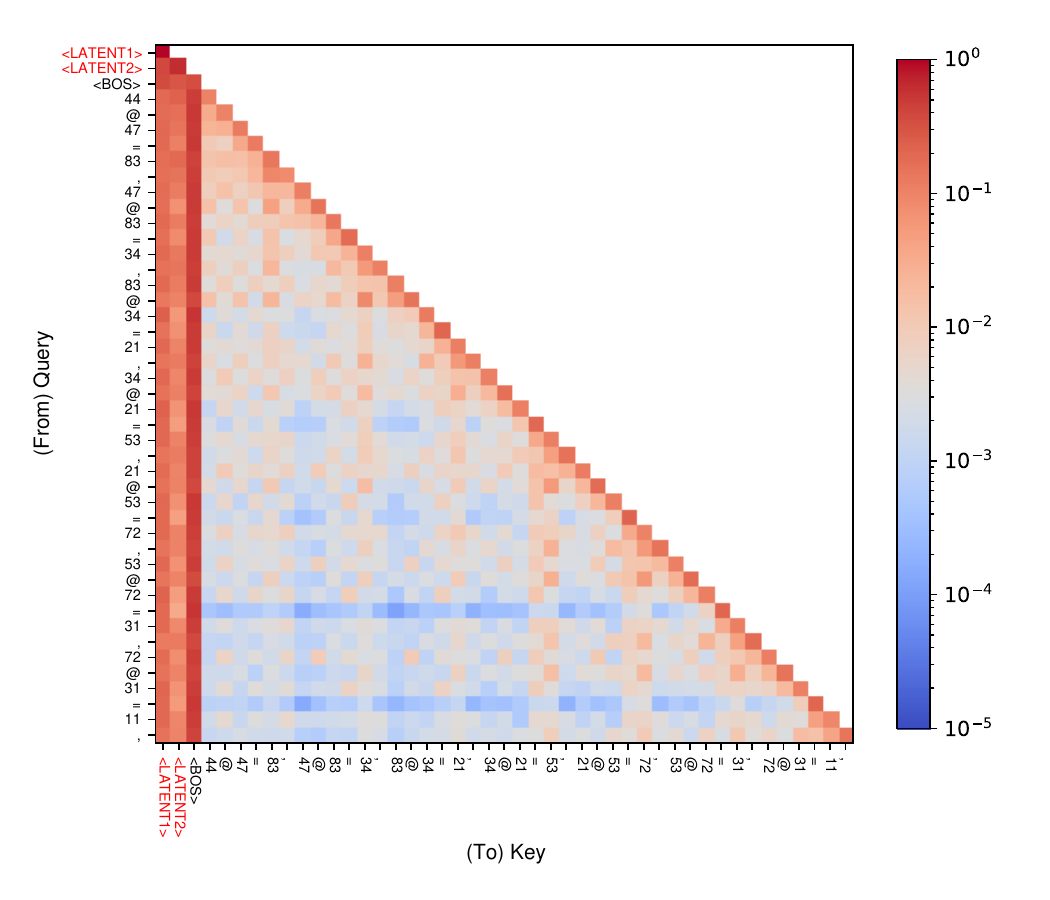}
        \caption{The last layer}
    \end{subfigure}
    \caption{Attention maps for \start{2}.}
    \label{fig:attention_map_prompt}
\end{figure*}

\begin{figure*}[htbp]
    \centering
    \begin{subfigure}{0.48\textwidth}
        \centering
        \includegraphics[width=\textwidth]{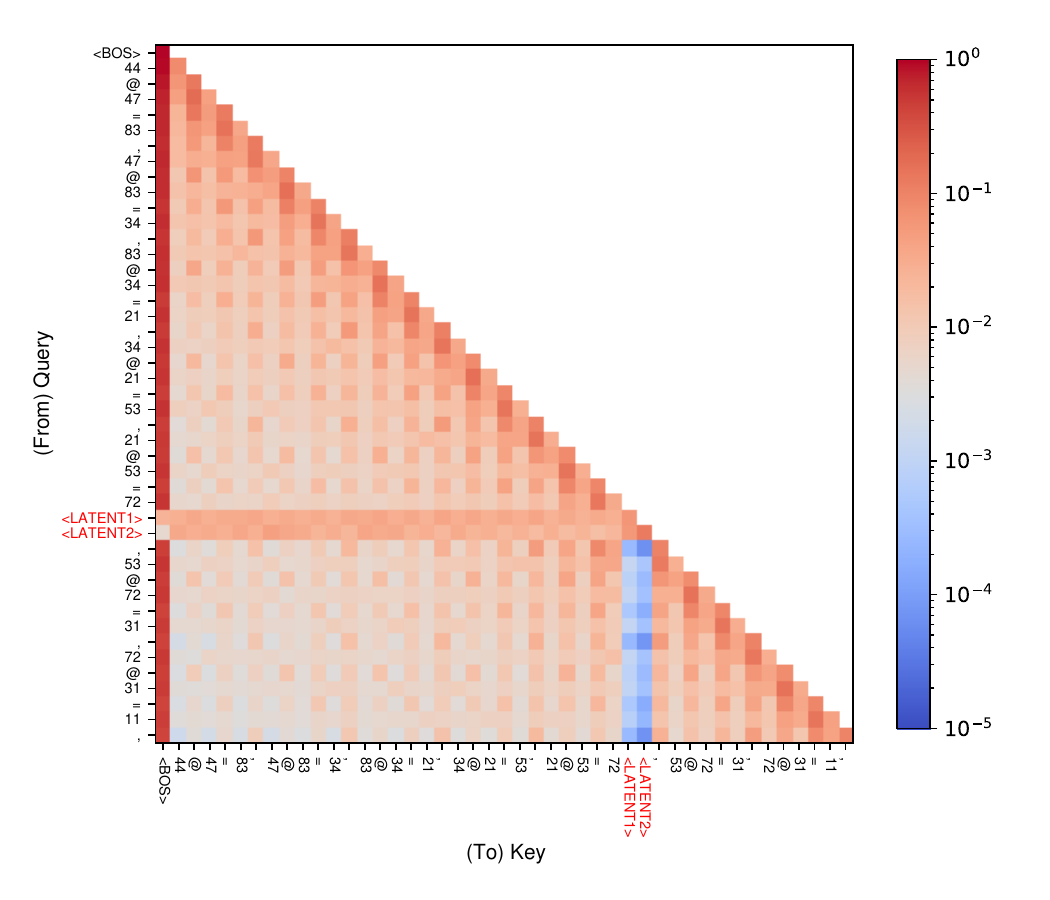}
        \caption{The first layer}
    \end{subfigure}
    \begin{subfigure}{0.48\textwidth}
        \centering
        \includegraphics[width=\textwidth]{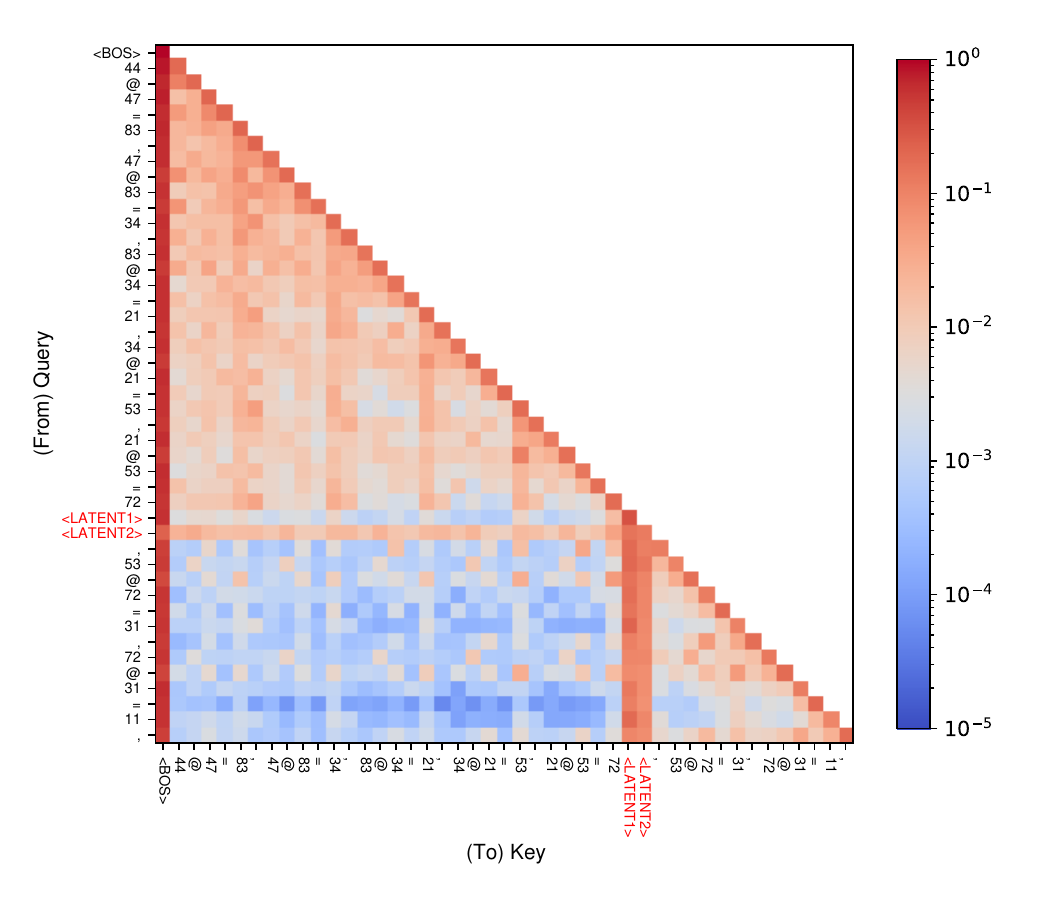}
        \caption{The last layer}
    \end{subfigure}
    \caption{Attention maps for \eend{2}.}
    \label{fig:attention_map_end}
\end{figure*}

We show the attention maps for the \start{2} method and the \eend{2} method in Figs.~\ref{fig:attention_map_prompt} and~\ref{fig:attention_map_end}, respectively.

\subsection{Attention Maps for the Repetition Task}\label{sec:attention_map_repetition}
\begin{figure*}[htbp]
    \centering
    \begin{subfigure}{0.48\textwidth}
        \centering
        \includegraphics[width=\textwidth]{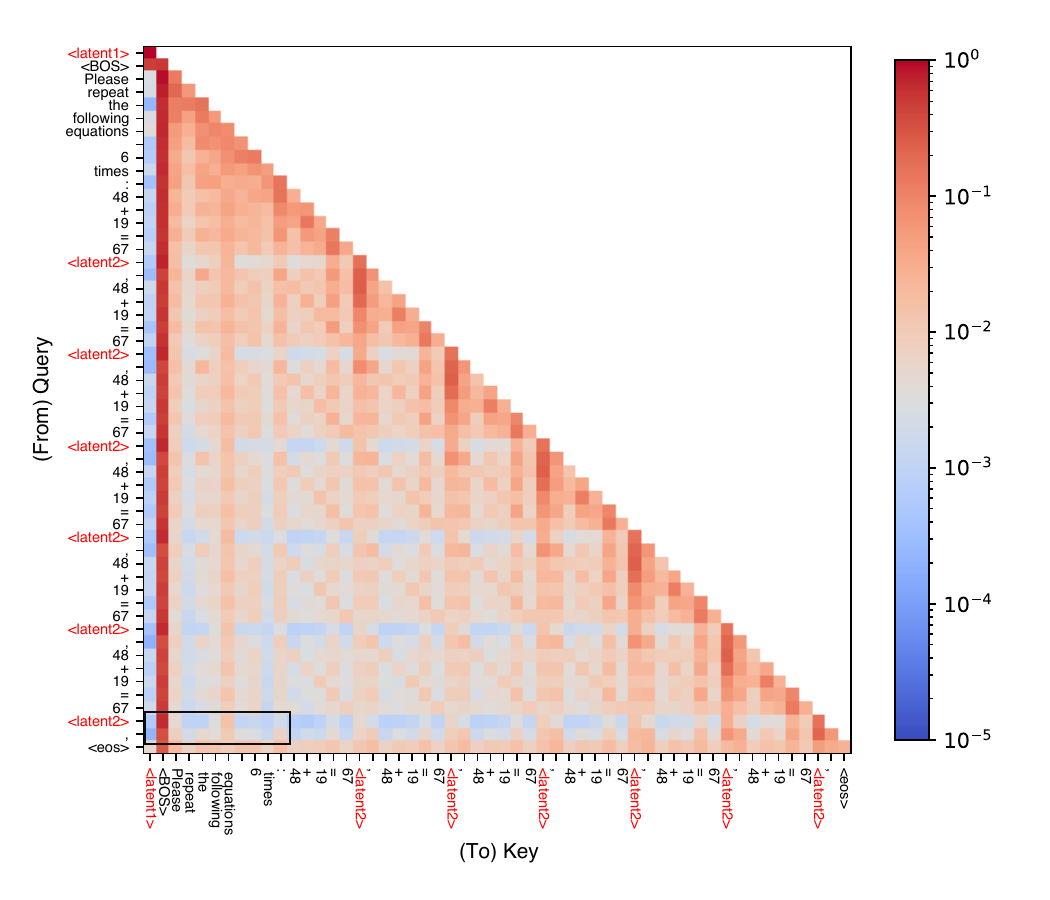}
        \caption{The first layer}
    \end{subfigure}
    \begin{subfigure}{0.48\textwidth}
        \centering
        \includegraphics[width=\textwidth]{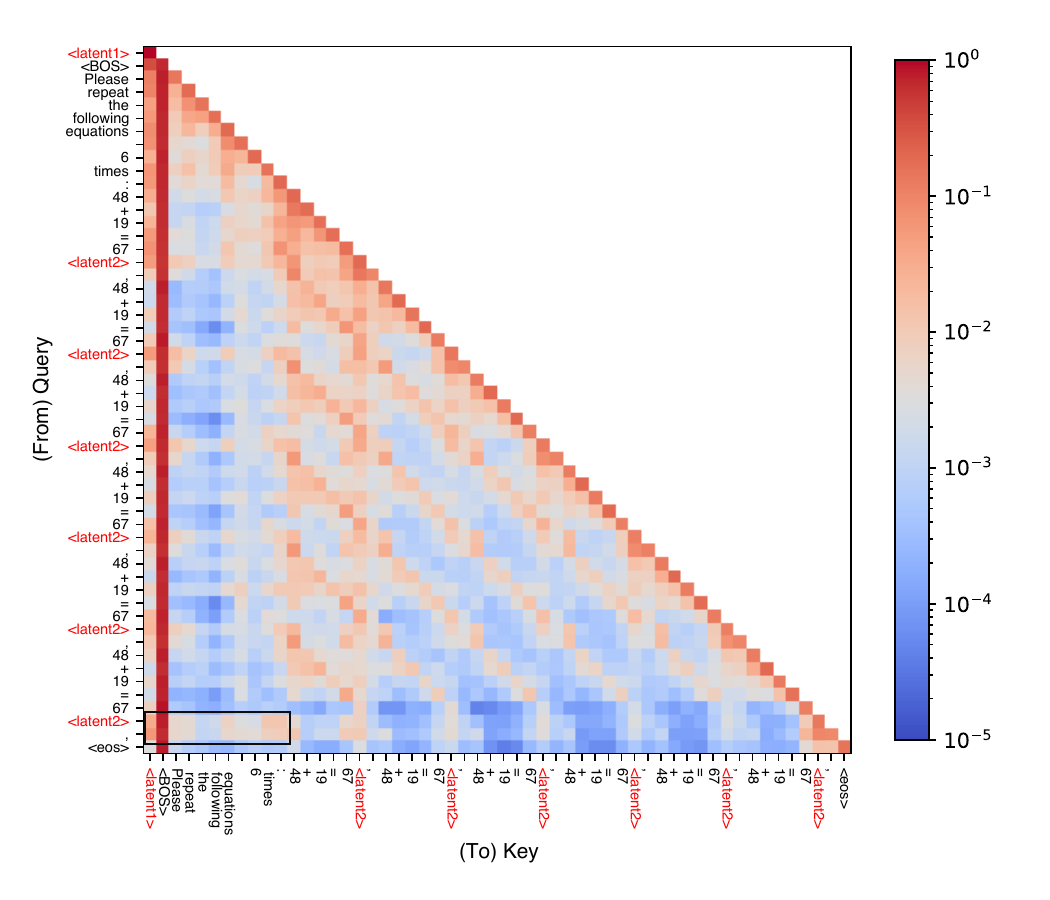}
        \caption{The last layer}
    \end{subfigure}
    \caption{Attention maps for \comma{1} (w/ \fs).}
    \label{fig:attention_map_repetition}
\end{figure*}

We show the attention maps for the proposed \comma{1} (with \fs) method in Fig.~\ref{fig:attention_map_repetition}.
We observe that in the last layer, the latent token and comma before the \eos token attend to the instruction tokens (i.e., ``Please repeat the following equation 6 times'') more strongly than to other tokens. This is some preliminary evidence that the latent tokens may have learned to recognize an ending signal for this task.

\subsection{Ablation Studies}\label{sec:ablation_synthetic}
We conduct the ablation studies on the synthetic tasks by evaluating the variants of \comma{2} with \llamas, as shown in Table~\ref{table:ablation_synthetic}. In particular, we concentrate on the impacts of freezing position IDs and prepend/append latent tokens on the model performance, namely, four setups are considered: (1) freeze \texttt{POS\_ID} \& \prepend (default setup), (2) freeze \texttt{POS\_ID} \& \append, (3) increase \texttt{POS\_ID} \& \prepend, and (4) increase \texttt{POS\_ID} \& \append.

While increasing position IDs significantly degrades the model performance, the design of appending tokens sometimes achieves better performance compared with other methods, e.g., in the Generation task.

\begin{table}[htbp]
\caption{Ablation studies on synthetic tasks, averaged over three random runs. We report the number of equations for the Generation task and the accuracy for the Summation task.}
\label{table:ablation_synthetic}
\centering
\begin{tabular}{l|cc|cc}
\toprule
& \multicolumn{2}{c|}{Generation} & \multicolumn{2}{c}{Summation (\%)} \\ 
\cmidrule(lr){2-3} \cmidrule(lr){4-5}
& 50 & 60 & (30, 50) & (50, 70) \\ \midrule
Freeze \texttt{POS\_ID} \& \prepend  (Default)    & 38.18{\footnotesize $\pm$6.69} & 40.46{\footnotesize $\pm$8.12} & \textbf{93.82}{\footnotesize $\pm$4.48} & \underline{72.40}{\footnotesize $\pm$5.31} \\
Freeze \texttt{POS\_ID} \& \append       & \textbf{41.83}{\footnotesize $\pm$2.42} & \textbf{44.50}{\footnotesize $\pm$4.07} & 88.67{\footnotesize $\pm$0.97} & 67.38{\footnotesize $\pm$5.98} \\
Increase \texttt{POS\_ID} \& \prepend & 1.34{\footnotesize $\pm$2.96} & 1.69{\footnotesize $\pm$2.96} & 88.09{\footnotesize $\pm$9.98} & 71.94{\footnotesize $\pm$20.29} \\
Increase \texttt{POS\_ID}  \& \append    & \underline{41.18}{\footnotesize $\pm$5.85} & \underline{43.65}{\footnotesize $\pm$9.74} & \underline{91.15}{\footnotesize $\pm$2.37} & \textbf{73.95}{\footnotesize $\pm$6.91} \\
\bottomrule
\end{tabular}
\end{table}

\clearpage
\section{Benchmark Evaluation}\label{sec:benchmark}
In this section, we conduct extensive experiments across a diverse range of benchmarks to demonstrate the improvements brought by latent tokens in different downstream tasks, showing their potential in broad applications.

\subsection{Setup}

\paragraph{Models \& Datasets.}
We evaluate the downstream task performance of the proposed method using the pre-trained \llamas base model\footnote{\url{https://huggingface.co/meta-llama/Llama-3.2-1B}} and \llamal base model\footnote{\url{https://huggingface.co/meta-llama/Llama-3.1-8B}}~\cite{meta2024llama3herdmodels}.
We adopt several benchmarks, including WikiSplit~\cite{wikisplit}, NarrativeQA~\cite{narrativeqa}, and GSM8K~\cite{gsm8k}, for comprehensively evaluating the language modeling, reading comprehension, and arithmetic reasoning capabilities of the LLMs.
The concrete examples for these datasets are referred to Table~\ref{table:examples_benchmark}.

\begin{itemize}
\item{WikiSplit~\cite{wikisplit}:} This dataset was constructed automatically from the publicly available Wikipedia revision history. Each sample contains an English sentence extracted from Wikipedia, to be split into simple sentences that preserve the original meaning. This dataset can be accessed via \url{https://huggingface.co/datasets/google-research-datasets/wiki_split}. We report the BLEU score for this dataset by comparing the generated response with the ground-truth answer.

\item{NarrativeQA~\cite{narrativeqa}:} This dataset is an English dataset of stories and corresponding questions designed to test reading comprehension. We format each prompt with its question and summary of the story. At the inference stage, each sample may have more than one provided answer, all of which are viewed as correct. This dataset can be accessed via \url{https://huggingface.co/datasets/deepmind/narrativeqa}. We report the Rouge-L score for this dataset by comparing the generated response with the ground-truth answer.

\item{GSM8K~\cite{gsm8k}:} This is a dataset of high-quality grade school math problems created by human problem writers. It is a frequently used benchmark to evaluate the arithmetic reasoning ability of LLM. This dataset can be accessed via \url{https://huggingface.co/datasets/openai/gsm8k}. Each training sample consists of a grade-school math problem, a step-by-step reasoning process, and a final answer. Both the reasoning process and the final answer are incorporated into the response and contribute to the training loss. To enhance the quality of the response, we use the CoT prompt ``Please think step by step.'' in the query. For evaluation, we calculate the accuracy by extracting the final answers from the responses.
\end{itemize}

\paragraph{Baselines \& Our method.}
Following the approaches in prompt tuning~\cite{prompt} and pause/filler tokens~\cite{google2024think,pfau2024lets}, we add $m$ latent tokens at the start (\start{m}) or end (\eend{m}) of the query, respectively.
For our method, we insert $m$ latent tokens every $k$ verbal tokens (\freq{k}{m}), with or without function specialization (\fs).

We run all experiments with three random seeds and report the mean and standard deviation of the results. In each setup (specified by the number of trainable parameters), the best results are highlighted in \textbf{bold}, while the second-best ones are \underline{underlined}.
The values of adopted hyperparameters are summarized in Table~\ref{table:hyperparameters}.

\begin{table*}[ht]
    \centering
    \caption{Results on the benchmark datasets. We report the BLEU score for the WikiSplit dataset, Rouge-L score for NarrativeQA dataset, and the accuracy for the GSM8K dataset. The base model is evaluated with few-shot prompting as a reference point.
    \start{m} and \eend{m} refer to adding $m$ latent tokens at the start or end of the query, respectively; \freq{k}{m} inserts $m$ latent tokens every $k$ verbal tokens; \fs refers to function specialization.} 
    \label{tab:main_results}
    \resizebox{\textwidth}{!}{
    \begin{tabular}{l|c|ccc|ccc}
    \toprule
     & & \multicolumn{3}{c|}{\llamas} & \multicolumn{3}{c}{\llamal} \\ 
    \cmidrule(lr){3-5} \cmidrule(lr){6-8}
     & \# of Params. & WikiSplit & NarrativeQA  & GSM8K (\%)    & WikiSplit & NarrativeQA    & GSM8K (\%)  \\
    \midrule
    Base Model & N/A     &  0.1517  &  0.0414 &   6.90   &  0.1359        & 0.0629   &    49.20   \\
    
    \midrule
    
    \start{1}  & 1$\times$ & \underline{0.6548}  {\footnotesize $\pm 0.0115$}  & \underline{0.5722} {\footnotesize $\pm 0.0357$}  & \underline{9.33} {\footnotesize $\pm 0.33$}  & \underline{0.7395} {\footnotesize $\pm 0.0157$}  & \underline{0.7494} {\footnotesize $\pm 0.0151$}  & \textbf{51.53} {\footnotesize $\pm 1.50$}  \\
    \eend{1}   & 1$\times$  & 0.6268 {\footnotesize $\pm 0.0072$}  & 0.4498 {\footnotesize $\pm 0.0289$}  & 8.21 {\footnotesize $\pm 0.61$}  & 0.7104{\footnotesize $\pm 0.0042$}  & 0.7070 {\footnotesize $\pm 0.0010$}  & 48.09 {\footnotesize $\pm 1.07$}  \\
    \rowcolor{cyan!15}
    \freq{8}{1}  & 1$\times$   & \textbf{0.6586} {\footnotesize $\pm 0.0008$}  & \textbf{0.6119} {\footnotesize $\pm 0.0103$}  & \textbf{9.60} {\footnotesize $\pm 0.43$}  & \textbf{0.7550} {\footnotesize $\pm 0.0040$}  & \textbf{0.7512} {\footnotesize $\pm 0.0143$} & \underline{51.45} {\footnotesize $\pm 1.27$}  \\
    
    \midrule
    
    \start{4}   & 4$\times$  & 0.7124 {\footnotesize $\pm 0.0079$}  & 0.6266 {\footnotesize $\pm 0.0201$}  & 10.67 {\footnotesize $\pm 0.46$}  & \underline{0.7824} {\footnotesize $\pm 0.0137$}  & 0.7557 {\footnotesize $\pm 0.0033$}  & 52.16 {\footnotesize $\pm 0.55$}  \\
    \eend{4}    & 4$\times$  & 0.6712 {\footnotesize $\pm 0.0031$}  & 0.4950 {\footnotesize $\pm 0.0055$}  & 11.07 {\footnotesize $\pm 0.66$}  & 0.7115 {\footnotesize $\pm 0.0371$}  & 0.7099 {\footnotesize $\pm 0.0035$}  & 50.01 {\footnotesize $\pm 0.70$}  \\
    \rowcolor{cyan!15}
    \freq{8}{4}    & 4$\times$    & \textbf{0.7200} {\footnotesize $\pm 0.0046$}  & \underline{0.6312} {\footnotesize $\pm 0.0056$}  & \underline{11.14} {\footnotesize $\pm 0.27$}  & 0.7720 {\footnotesize $\pm 0.0105$}  & \textbf{0.7672} {\footnotesize $\pm 0.0046$}   & \textbf{53.73} {\footnotesize $\pm 1.67$}  \\
    \rowcolor{cyan!15}
    \freq{8}{1} (w/ \fs)  & 4$\times$ & \underline{0.7178} {\footnotesize $\pm 0.0070$}  & \textbf{0.6389} {\footnotesize $\pm 0.0084$}  & \textbf{12.46} {\footnotesize $\pm 0.37$}  & \textbf{0.7855} {\footnotesize $\pm 0.0056$}  & \underline{0.7606} {\footnotesize $\pm 0.0086$}  & \underline{52.57} {\footnotesize $\pm 1.48$}  \\

    \midrule

    \start{16} & 16$\times$ & 0.7605 {\footnotesize $\pm 0.0028$}  & 0.6557 {\footnotesize $\pm 0.0078$}  & 13.52 {\footnotesize $\pm 0.86$}  & \textbf{0.8001} {\footnotesize $\pm 0.0116$}  & \underline{0.7665} {\footnotesize $\pm 0.0099$}  & 54.36 {\footnotesize $\pm 0.80$}  \\
    \eend{16}  & 16$\times$  & 0.7042 {\footnotesize $\pm 0.0036$}  & 0.5295 {\footnotesize $\pm 0.0105$}  & 12.64 {\footnotesize $\pm 0.27$}  & 0.7652 {\footnotesize $\pm 0.0019$}  & 0.7261 {\footnotesize $\pm 0.0101$}  & 50.01 {\footnotesize $\pm 1.23$}  \\
    \rowcolor{cyan!15}
    \freq{8}{4} (w/ \fs) & 16$\times$ & \underline{0.7619} {\footnotesize $\pm 0.0049$}  & \underline{0.6597} {\footnotesize $\pm 0.0138$}  & \underline{13.95} {\footnotesize $\pm 0.33$}  & 0.7813 {\footnotesize $\pm 0.0257$}  & 0.7642 {\footnotesize $\pm 0.0103$}  & \textbf{55.30} {\footnotesize $\pm 1.89$}  \\
    \rowcolor{cyan!15}
    \freq{32}{4} (w/ \fs) & 16$\times$ & \textbf{0.7647} {\footnotesize $\pm 0.0245$}  & 0.6555 {\footnotesize $\pm 0.0019$}  & \textbf{14.25} {\footnotesize $\pm 1.10$}  & \underline{0.7963} {\footnotesize $\pm 0.0032$}  & 0.7648 {\footnotesize $\pm 0.0020$}  & \underline{54.79} {\footnotesize $\pm 0.65$}  \\
    \rowcolor{cyan!15}
    \freq{64}{4} (w/ \fs) & 16$\times$ & 0.7568 {\footnotesize $\pm 0.0045$}  & \textbf{0.6651} {\footnotesize $\pm 0.0027$}  & 12.31 {\footnotesize $\pm 0.95$} & 0.7952 {\footnotesize $\pm 0.0030$}  & \textbf{0.7702} {\footnotesize $\pm 0.0012$}  & 53.43 {\footnotesize $\pm 1.15$}  \\

    \bottomrule
    \end{tabular}
    }
\end{table*}

\subsection{Performance Comparison}
Table \ref{tab:main_results} shows the results of various methods with different numbers of trainable parameters by controlling the number of latent tokens. 
We observe that the proposed method, i.e., integrating latent tokens within the sequence periodically, achieves better performances than the baselines in most cases, especially when fine-tuning the \llamas model.
In comparison, adding special tokens between the query and generated response underperforms other methods, but still outperforms the naive baseline (Base Model) significantly.
These results demonstrate that the proposed approach of inserting latent tokens successfully enhances the intermediate computation in the inference process, thereby improving the quality of responses.

Furthermore, we notice that the proposed method \freq{8}{4} has similar amounts of computation per sequence with \freq{8}{4} (w/ \fs). Nevertheless, the latter obtains a large improvement by specializing the tokens in different functions. A similar phenomenon is observed when comparing \freq{8}{1} and \freq{32}{4} (w/ \fs). Without \fs, the same latent tokens may have to learn conflicting objectives for different purposes, leading to slightly worse performance.
These results highlight the effectiveness of the proposed function specialization design in solving complex problems.\looseness=-1

\subsection{Ablation Studies}\label{sec:ablation}
We conduct the ablation study by inserting latent tokens with the \freq{8}{4} approach with \llamas, as shown in Table~\ref{table:ablation_benchmark}. 
Similar to Appendix~\ref{sec:ablation_synthetic}, we consider four setups: (1) freeze \texttt{POS\_ID} \& \prepend (default setup), (2) freeze \texttt{POS\_ID} \& \append, (3) increase \texttt{POS\_ID} \& \prepend, and (4) increase \texttt{POS\_ID} \& \append.
We observe that adopting the fixed position IDs effectively stabilizes the training and avoids bad performance; the design of prepending latent tokens also enhances the training performance.

\begin{table}[htbp]
\caption{Ablation studies on benchmark datasets.}
\label{table:ablation_benchmark}
\centering
\begin{tabular}{l|ccc}
\toprule
& WikiSplit & NarrativeQA & GSM8K (\%)  \\ \midrule
Freeze \texttt{POS\_ID} \& \prepend (Default)    & \textbf{0.7200}{\footnotesize $\pm$ 0.0046}  & \textbf{0.6312}{\footnotesize $\pm$0.0056}  & \textbf{11.14}{\footnotesize $\pm$0.27} \\ 
Freeze \texttt{POS\_ID} \& \append  &    0.7146{\footnotesize $\pm$0.0081}      & \underline{0.6239}{\footnotesize $\pm$0.0086}    &  \underline{10.71}{\footnotesize $\pm$0.39}    \\
Increase \texttt{POS\_ID} \& \prepend &  0.6902{\footnotesize $\pm$0.0086}         &  0.6174{\footnotesize $\pm$0.0093}   &  9.88{\footnotesize $\pm$1.73}        \\
Increase \texttt{POS\_ID}  \& \append     &   \underline{0.7122}{\footnotesize $\pm$0.0053}     &   0.6109{\footnotesize $\pm$0.0121}  & 10.59{\footnotesize $\pm$0.90}    \\ 
\bottomrule
\end{tabular}
\end{table}

\subsection{Comparison with Pause Tokens}\label{sec:further_comparison}

We compare the proposed method \freq{8}{4} (w/ \fs) with pause tokens~\cite{google2024think} by adding 16 \pause tokens, with the same number of parameters as \freq{8}{4} (w/ \fs), at the end of the query with the \append option. The results in Table~\ref{table:compare_pause} demonstrate the superior performance of our proposed method thanks to the dedicated designs.

\begin{table}[htbp]
\caption{Comparisons with pause token.}
\label{table:compare_pause}
\centering
\begin{tabular}{l|l|ccc}
\toprule
\multicolumn{2}{c|}{} & WikiSplit & NarrativeQA & GSM8K (\%)  \\ \midrule

\multirow{2}{*}{\llamas} & Pause Token &  0.6975{\footnotesize $\pm 0.0035$}  & 0.5090{\footnotesize $\pm 0.0141$}    & 12.33{\footnotesize $\pm 0.52$}  \\
& \cellcolor{cyan!15}Ours   & \cellcolor{cyan!15}\textbf{0.7619}{\footnotesize $\pm0.0049$}  & \cellcolor{cyan!15}\textbf{0.6597}{\footnotesize $\pm0.0138$}  & \cellcolor{cyan!15}\textbf{13.95}{\footnotesize $\pm 0.33$} \\  \midrule

\multirow{2}{*}{\llamal} & Pause Token & 0.7599{\footnotesize $\pm 0.0006$}  & 0.7160{\footnotesize $\pm 0.0033$}  & 49.78{\footnotesize $\pm 1.61$} \\
& \cellcolor{cyan!15}Ours   & \cellcolor{cyan!15}\textbf{0.7961}{\footnotesize $\pm 0.0028$}  & \cellcolor{cyan!15}\textbf{0.7642}{\footnotesize $\pm 0.0103$}  & \cellcolor{cyan!15}\textbf{55.30}{\footnotesize $\pm 1.89$} \\
\bottomrule
\end{tabular}
\end{table}

\begingroup
\setlength{\tabcolsep}{3.5pt}
\begin{table*}[t]
    \caption{Examples of benchmark datasets. \textbf{\textbackslash n} is the delimiter.}
    \label{table:examples_benchmark}
    \centering
    \begin{tabular}{l p{11.0cm}}
        \toprule
        Dataset & Example \\ \midrule
        WikiSplit & \query Split the sentence below into separate sentences.\textbf{\textbackslash n\textbackslash n}Etchingham was born and grew up in Leicester where her parents were teachers, and got her first job in journalism with BBC Radio Leicester whilst still at school.\textbf{\textbackslash n\textbackslash n}\\
        & \response Etchingham was born. Etchingham grew up in Leicester. Her parents were teachers. Etchingham got her first job in journalism with BBC Radio Leicester whilst still at school.\\  \midrule 
        
        NarrativeQA & \query Please answer the question based on the context.\textbf{\textbackslash n\textbackslash n}Question: Who does echo weep?\textbf{\textbackslash n}Context:  The play begins with three pages disputing over the black cloak usually worn by the actor who delivers the prologue. They draw lots for the cloak, and one of the losers, Anaides, starts telling the audience what happens in the play to come; the others try to suppress him, interrupting him and putting their hands over his mouth. Soon they are fighting over the cloak and criticizing the author and the spectators as well.\textbf{\textbackslash n}In the play proper, the goddess Diana, also called Cynthia, has ordained a "solemn revels" in the valley of Gargaphie in Greece. The gods Cupid and Mercury appear, and they too start to argue. Mercury has awakened Echo, who weeps for Narcissus, and states that a drink from Narcissus\'s spring causes the drinkers to "Grow dotingly enamored of themselves." The courtiers and ladies assembled for the Cynthia\'s revels all drink from the spring.\textbf{\textbackslash n}Asotus, a foolish spendthrift who longs to become a courtier and a master of fashion and manners, also drinks from the spring; emboldened by vanity and self-love, he challenges all comers to a competition of "court compliment." The competition is held, in four phases, and the courtiers are beaten. Two symbolic masques are performed within the play for the assembled revelers. At their conclusion, Cynthia (representing Queen Elizabeth) has the dancers unmask and shows that vices have masqueraded as virtues. She sentences them to make reparation and to purify themselves by bathing in the spring at Mount Helicon.\textbf{\textbackslash n}The figure of Actaeon in the play may represent Robert Devereux, 2nd Earl of Essex, while Cynthia\'s lady in waiting Arete may be Lucy, Countess of Bedford, one of Elizabeth\'s ladies in waiting as well as Jonson\'s patroness.\textbf{\textbackslash n}The play is notably rich in music, as is typical for the theatre of the boys\' companies, which originated as church choirs.\textbf{\textbackslash n}Answer:\\
        & \response Narcissus \\\midrule 
        
        GSM8K & \query Please answer the following question. Please think step by step.\textbf{\textbackslash n\textbackslash n}Question: James writes a 3-page letter to 2 different friends twice a week. How many pages does he write a year? \textbf{\textbackslash n\textbackslash n}Answer: \\
        & \response He writes each friend 3*2=6 pages a week So he writes 6*2=12 pages every week That means he writes 12*52=624 pages a year \#\#\#\# 624 \\ \bottomrule
    \end{tabular}
\end{table*}
\endgroup

\begin{table}[htbp]
    \caption{Experimental setup for benchmark datasets.}
    \label{table:hyperparameters}
    \centering
    \begin{tabular}{l|ccc}
    \toprule
    Dataset                          & WikiSplit & NarrativeQA & GSM8K \\ \midrule
    \multicolumn{4}{c}{\textbf{Training}} \\ \midrule
    Optimizer                        & \multicolumn{3}{c}{AdamW}     \\ 
    Weight Decay                     &  \multicolumn{3}{c}{1e-4}   \\ 
    
    Warmup Ratio                     &  \multicolumn{3}{c}{0.01}          \\ 
    Learning Rate Schedule           &  \multicolumn{3}{c}{Cosine}        \\ 
    Learning Rate                    &  \multicolumn{3}{c}{5e-3}       \\
    Minimal Learning Rate                    &  \multicolumn{3}{c}{5e-5}       \\
    Batch Size (\llamas)                      & 32 & 16 & 32             \\
    Batch Size (\llamal)          & 16 & 16 & 32             \\
    \# of Epochs                &  1 & 1 & 30         \\  
    \# of Training Samples & 154,582 & 32,747 & 5,976 \\
    \# of Validation Samples & 2,048 & 3,461 & 1,496 \\  \midrule
    \multicolumn{4}{c}{\textbf{Inference}} \\ \midrule
    \# of Test Samples & 1,024 & 1,024 & 1,319 \\ 
    Maximal \# of New Tokens     & 200 & 50  & 200       \\
    \# of Shots for Evaluating Base Model & 5 & 3 & 8$^*$ \\
    \bottomrule
    \multicolumn{4}{l}{%
    \footnotesize \parbox{10cm}{%
    $^*$The few-shot examples for GSM8K are the same as those in~\cite{cot}.}}
    \end{tabular}
\end{table}

\end{document}